\documentclass[10pt,twocolumn,letterpaper]{article}

\usepackage[pagenumbers]{cvpr} 

\usepackage{enumitem}
\newcommand\mypara[1]{\vspace{1.2mm}\noindent\textbf{#1}}
\definecolor{cvprblue}{rgb}{0.21,0.49,0.74}
\usepackage[pagebackref,breaklinks,colorlinks,allcolors=cvprblue]{hyperref}
\usepackage{comment}
\usepackage{dblfloatfix}

\usepackage{amssymb}
\usepackage{amsmath,amsfonts}
\usepackage{amsopn}
\usepackage{bm} %
\usepackage{multirow}
\usepackage{tabularx}
\newcommand{\vct}[1]{\boldsymbol{#1}} %
\newcommand{\mat}[1]{\boldsymbol{#1}} %

\newcommand{\OursD}{\method{Prompt-CAM-Deep}\xspace}
\newcommand{\OursS}{\method{Prompt-CAM-Shallow}\xspace}
\newcommand{\Ours}{\method{Prompt-CAM}\xspace}

\newcommand{\field}[1]{\mathbb{#1}}

\newcommand{\R}{\field{R}} %

\newcommand{\ProbOpr}[1]{\mathbb{#1}}

\newcommand{\expect}[2]{%
\ifthenelse{\equal{#2}{}}{\ProbOpr{E}_{#1}}
{\ifthenelse{\equal{#1}{}}{\ProbOpr{E}\left[#2\right]}{\ProbOpr{E}_{#1}\left[#2\right]}}} %
\newcommand{\var}[2]{%
\ifthenelse{\equal{#2}{}}{\ProbOpr{VAR}_{#1}}
{\ifthenelse{\equal{#1}{}}{\ProbOpr{VAR}\left[#2\right]}{\ProbOpr{VAR}_{#1}\left[#2\right]}}} %

\DeclareMathOperator{\argmax}{arg\,max}

\DeclareMathOperator{\softmax}{softmax}

\newcommand{\vp}{\vct{p}}
\newcommand{\vq}{\vct{q}}
\newcommand{\ve}{{\vct{e}}}
\newcommand{\vx}{{\vct{x}}}
\newcommand{\valpha}{{\vct{\alpha}}}

\newcommand{\vz}{{\vct{z}}}

\newcommand{\vv}{\vct{v}}
\newcommand{\vw}{\vct{w}}

\newcommand{\mP}{\mat{P}}

\newcommand{\mE}{\mat{E}}

\newcommand{\mV}{\mat{V}}

\newcommand{\mZ}{\mat{Z}}

\newcommand{\mI}{\mat{I}}
\newcommand{\mK}{\mat{K}}

\newcommand{\method}[1]{\textsc{#1}}
\newcommand{\eat}[1]{}

\title{\Ours: Making Vision Transformers {Interpretable}\\ for Fine-Grained Analysis}

\author{Arpita Chowdhury\textsuperscript{1}, Dipanjyoti Paul\textsuperscript{2}, Zheda Mai\textsuperscript{1}, Jianyang Gu\textsuperscript{1}, Ziheng Zhang\textsuperscript{1},\\ Kazi Sajeed Mehrab\textsuperscript{3}, Elizabeth G. Campolongo\textsuperscript{1}, Daniel Rubenstein\textsuperscript{4}, Charles V. Stewart\textsuperscript{5},\\ Anuj Karpatne\textsuperscript{3}, Tanya Berger-Wolf\textsuperscript{1}, Yu Su\textsuperscript{1}, Wei-Lun Chao\textsuperscript{1}
\\
\vspace{-5pt}
\\ 
    \textsuperscript{1}The Ohio State University, \textsuperscript{2}University of Tsukuba, \textsuperscript{3}Virginia Tech, \textsuperscript{4}Princeton University,\\
    \textsuperscript{5}Rensselaer Polytechnic Institute
} 

\begin{document}

\twocolumn[{
\renewcommand\twocolumn[1][]{#1}
\maketitle
\begin{center}
    \centering
    \includegraphics[width=1\linewidth]{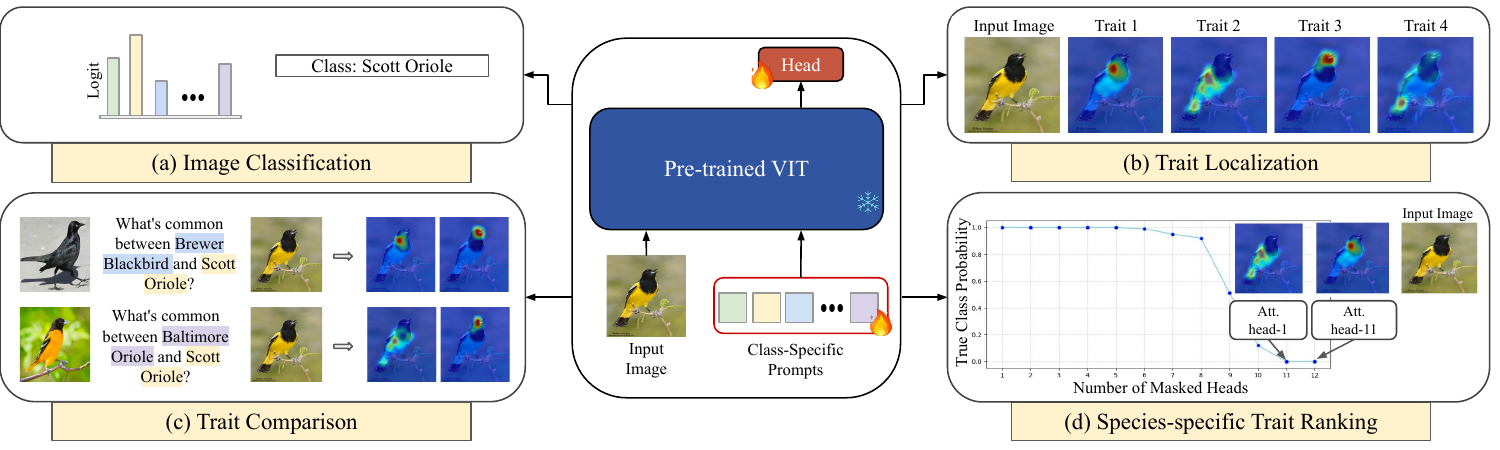}
    \vskip-5pt
    \captionof{figure}{\textbf{Illustration of \Ours.} By learning class-specific prompts for a pre-trained Vision Transformer (ViT), \Ours enables multiple functionalities. (a) \Ours achieves fine-grained image classification using the output logits from the class-specific prompts. (b) \Ours enables trait localization by visualizing the multi-head attention maps queried by the true-class prompt. (c) \Ours identifies common traits shared between species by visualizing the attention maps queried by another-class prompt. (d) \Ours can identify the most discriminative traits per species (\eg, distinctive yellow chest and black neck for ``Scott Oriole'') by systematically masking out the least important attention heads. See \autoref{ss:vis} for details.
    }
    \label{fig:teaser}
\end{center}
}]

\begin{abstract}
We present a simple approach to make pre-trained Vision Transformers (ViTs) interpretable for fine-grained analysis, aiming to identify and localize the traits that distinguish visually similar categories, such as bird species. Pre-trained ViTs, such as DINO, have demonstrated remarkable capabilities in extracting localized, discriminative features. However, saliency maps like Grad-CAM often fail to identify these traits, producing blurred, coarse heatmaps that highlight entire objects instead. We propose a novel approach, \textbf{Prompt Class Attention Map (\Ours)}, to address this limitation.  \Ours learns class-specific prompts for a pre-trained ViT and uses the corresponding outputs for classification. To correctly classify an image, the true-class prompt must attend to unique image patches not present in other classes' images  (\ie, traits). As a result, the true class's multi-head attention maps reveal traits and their locations. Implementation-wise, \Ours is almost a ``free lunch,'' requiring only a modification to the prediction head of Visual Prompt Tuning (VPT). This makes \Ours easy to train and apply, in stark contrast to other interpretable methods that require designing specific models and training processes. Extensive empirical studies on a dozen datasets from various domains (\eg, birds, fishes, insects, fungi, flowers, food, and cars) validate the superior interpretation capability of \Ours. The source code and demo are available at \url{https://github.com/Imageomics/Prompt_CAM}.
\end{abstract}

\section{Introduction}
\label{sec:intro}

Vision Transformers (ViT)~\cite{dosovitskiy2021an} pre-trained on huge datasets have greatly improved vision recognition, even for fine-grained objects~\cite{wang2023open,tang2023weakly, zhu2022dual, he2022transfg}. DINO \cite{caron2021emerging} and DINOv2 \cite{oquab2023dinov2} further showed remarkable abilities to extract features that are localized and informative, precisely representing the corresponding coordinates in the input image. These advancements open up the possibility of using pre-trained ViTs to discover ``traits'' that highlight each category's identity and distinguish it from other visually close ones.

One popular approach to this is saliency maps, for example, Class Activation Map (CAM)~\cite{zhou2016learning,selvaraju2017grad, muhammad2020eigen, jiang2021layercam}. After extracting the feature maps from an image, CAM highlights the spatial grids whose feature vectors align with the target class's fully connected weight. While easy to implement and efficient, the reported CAM saliency on ViTs is often far from expectation. It frequently locates the whole object with a blurred, coarse heatmap, instead of focusing on subtle traits that tell visually similar objects (\eg, birds) apart. One may argue that CAM was not originally developed for ViTs, but even with dedicated variants like attention rollout~\cite{kashefi2023explainability, chefer2021transformer, abnar2020quantifying}, the issue is only mildly attenuated.

\emph{What if we look at the attention maps?} ViTs rely on self-attention to relate image patches; the [CLS] token aggregates image features by attending to informative patches. As shown in~\cite{darcet2023vision, tang2023emergent, ng2023dreamcreature}, the attention maps of the [CLS] token do highlight local regions inside the object. \emph{However, these regions are not ``class-specific.''} Instead, they often focus on the same object regions across different categories, such as body parts like heads, wings, and tails of bird species. While these are where traits usually reside, they are not traits. For example, the distinction between ``Red-winged Blackbird'' and other bird species is the red spot on the wing, having little to do with other body parts.  
\begin{center}
\color{blue}
\emph{
How can we leverage pre-trained ViTs, particularly their localized and informative patch features, to identify traits that are so special for each category?
}
\end{center}

Our proposal is to \emph{prompt} ViTs with learnable ``class-specific'' tokens, one for each class, inspired by~\cite{paul2024simple,liu2021query2label,xu2022multi, li2023transcam}. These ``class-specific'' tokens, once inputted into ViTs, \emph{attend} to image patches via self-attention, similar to the [CLS] token. However, unlike the [CLS] token, which is ``class-agnostic,'' these ``class-specific'' tokens can \emph{attend to the same image differently}, with the potential to highlight regions specific to the corresponding classes, \ie, traits.

We implement our approach, named \textbf{Prompt Class Attention Map (\Ours)}, as follows. Given a pre-trained ViT and a fine-grained classification dataset with $C$ classes, we add $C$ learnable tokens as additional inputs alongside the input image. To make these tokens ``class-specific,'' we collect their corresponding output vectors after the final Transformer layer and perform inner products with a shared vector (also learnable) to obtain $C$ ``class-specific'' scores, following~\cite{paul2024simple}. One may interpret each class-specific score as how clearly the corresponding class's traits are visible in the input image. Intuitively, the input image's ground-truth class should possess the highest score, and we encourage this by minimizing a cross-entropy loss, treating the scores as logits. We keep the whole pre-trained ViT frozen and only optimize the $C$ tokens and the shared scoring vector. See \autoref{sec:method} for details and variants.

For interpretation during inference, we input the image and the $C$ tokens simultaneously to the ViT to obtain the $C$ scores. One can then select a specific class (\eg, the highest-score class) and visualize its multi-head attention maps over the image patches. See \autoref{fig:teaser} for an illustration and  \autoref{sec:method} for how to rank these maps to highlight the most discriminative traits. When the highest-score class is the ground-truth class, the attention maps reveal its traits. Otherwise, comparing the attention maps of the highest-score class with those of the ground-truth class helps explain why the image is misclassified. Possible reasons include the object being partially occluded or in an unusual pose, making its traits invisible, or the appearance being too similar to a wrong class, possibly due to lighting conditions (\autoref{fig:misclassified_images}).

\textbf{\Ours is fairly easy to implement and train.} 
\emph{It requires no change to pre-trained ViTs and no specially designed loss function or training strategy}---just the standard cross-entropy loss and SGD. Indeed, building upon Visual Prompt Tuning (VPT)~\cite{jia2022visual}, one merely needs to adjust a few lines of code and can enjoy fine-grained interpretation.
This simplicity sharply contrasts other interpretable methods like ProtoPNet~\cite{chen2019looks} and ProtoTree~\cite{nauta2021neural}. 
Compared to INterpretable TRansformer (INTR) \cite{paul2024simple}, which also featured simplicity, \Ours has three notable advantages.
First, \Ours is \emph{encoder-only} and can potentially utilize any ViT encoder. In contrast, INTR is built upon an encoder-decoder model pre-trained on object detection datasets. As a result, \Ours can more easily leverage up-to-date pre-trained models.  Second, \Ours can be trained much faster---only the prompts and the shared vector need to be learned. In contrast, INTR typically requires full fine-tuning. Third, \Ours produces cleaner and sharper attention maps than INTR, which we attribute to the use of state-of-the-art ViTs like DINO~\cite{caron2021emerging} or DINOv2~\cite{oquab2023dinov2}. Taken together, we view \Ours as a \emph{simpler} yet more powerful interpretable Transformer. 

\begin{figure}[!t]
\centering
\includegraphics[width=0.95\linewidth]{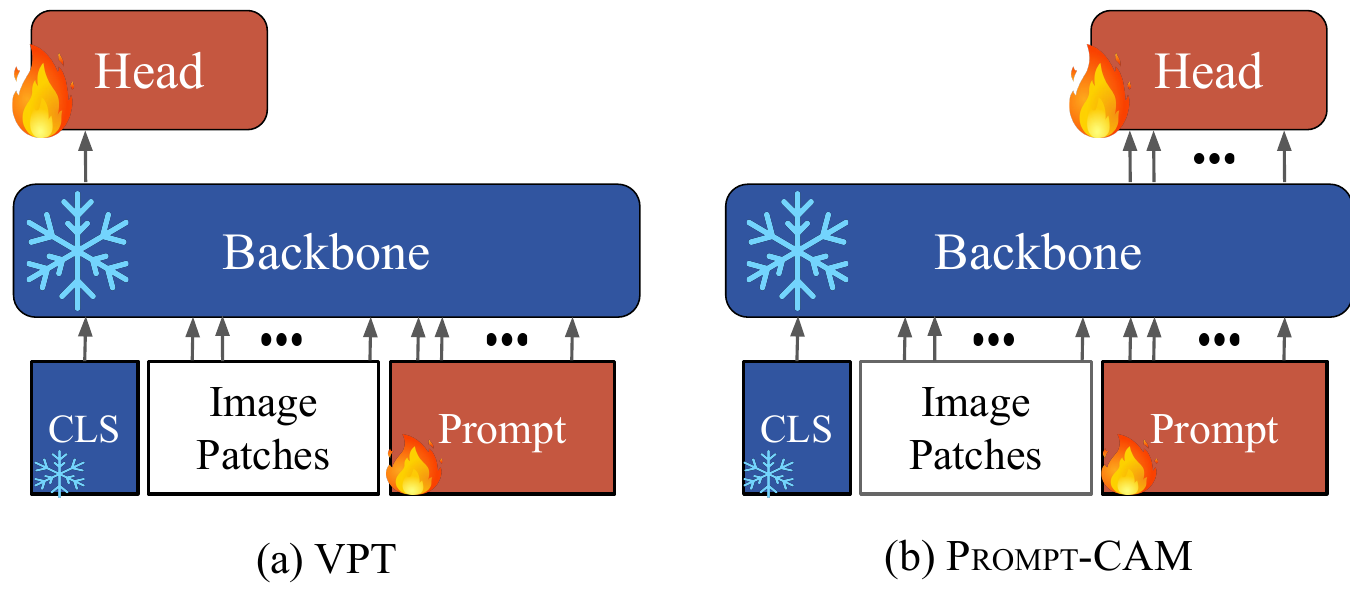}
\\
\vskip-7pt
\caption{\small \textbf{\Ours vs.~Visual Prompt Tuning (VPT)}. (a) VPT~\cite{jia2022visual} adds the prediction head on top of the [CLS] token's output, a default design to use ViTs for classification. (b) \Ours adds the prediction head on top of the injected prompts' outputs, making them class-specific to identify and localize traits.} 
\vskip-10pt
\label{fig-4: bird and spider}
\end{figure}

We validate \Ours on over a dozen datasets: CUB-200-2011~\cite{wah2011caltech}, Birds-525~\cite{piosenka2023birds}, Oxford
Pet~\cite{parkhi2012cats}, Stanford Dogs~\cite{khosla2011novel}, Stanford Cars~\cite{krause20133d}, iNaturalist-2021-Moths~\cite{van2021benchmarking}, Fish Vista~\cite{mehrab2024fish}, Rare Species~\cite{rare_species_dataset}, Insects-2~\cite{wu2019ip102}, iNaturalist-2021-Fungi~\cite{van2021benchmarking}, Oxford Flowers~\cite{nilsback2008automated}, Medicinal Leaf~\cite{roopashree2020medicinal}, Stanford Cars~\cite{krause20133d}, and Food 101~\cite{bossard2014food}. \Ours can identify different traits of a category through multi-head attention and consistently localize them in images. \emph{To our knowledge, \Ours is the only explainable or interpretable method for vision that has been evaluated on such a broad range of domains.} We further show \Ours's extendability by applying it to discovering taxonomy keys. Our contributions are two-fold. 
\begin{itemize}[nosep,topsep=1pt,parsep=0pt,partopsep=1pt, leftmargin=*]
\item We present \textbf{\Ours}, an easily implementable, trainable, and reproducible \emph{interpretable} method that leverages the representations of pre-trained ViTs to identify and localize traits for fine-grained analysis.
\item We conduct extensive experiments on more than a dozen datasets to validate \textbf{\Ours}'s interpretation quality, wide applicability, and extendability.  
\end{itemize}

\begin{figure*}[htpb!]
    \centering
    \includegraphics[width=0.92\linewidth]{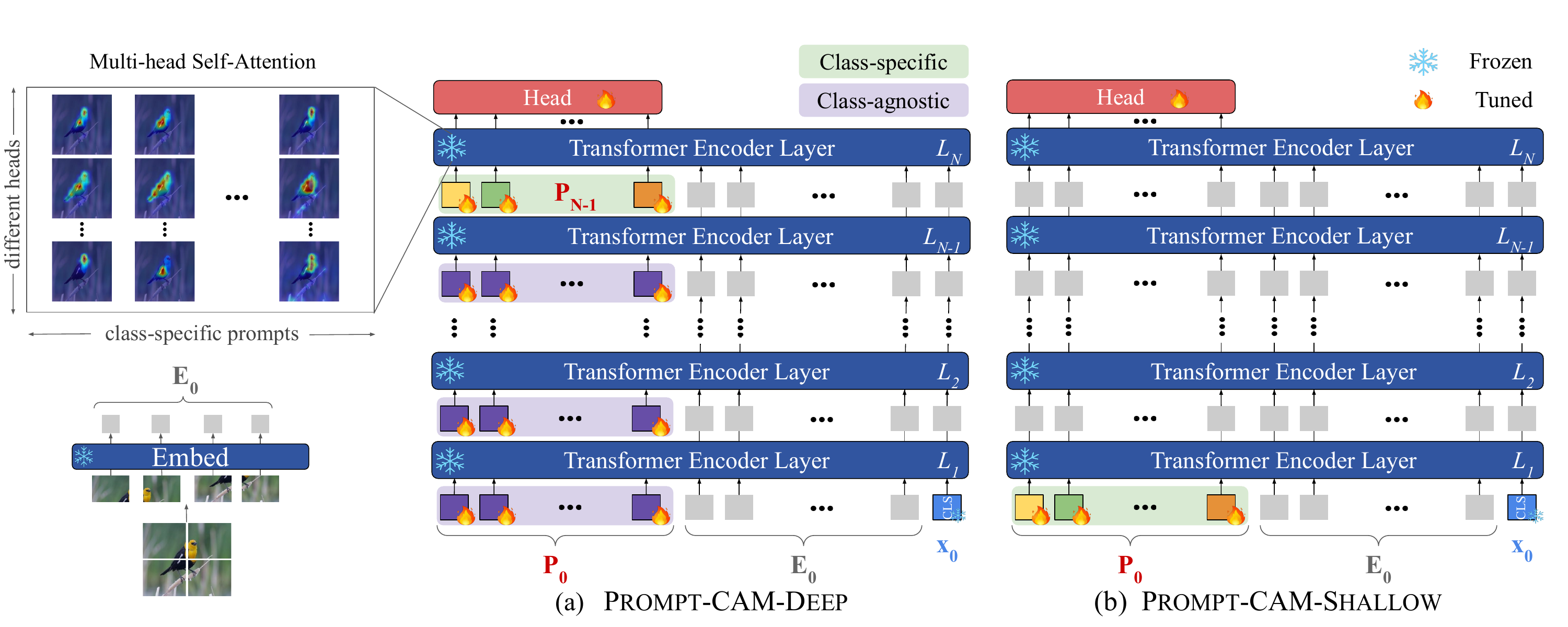}
    \vskip-8pt
    \caption{\small \textbf{Overview of Prompt Class Attention Map (\Ours)}. We explore two variants,  given a pre-trained ViT with $N$ layers and a downstream task with $C$ classes: (a) \OursD: insert $C$ learnable ``class-specific'' tokens to the \emph{last} layer's input and $C$ learnable ``class-agnostic'' tokens to each of the other $N-1$ layers' input; (b) \OursS: insert $C$ learnable ``class-specific'' tokens to the \emph{first} layer's input. During training, only the prompts and the prediction head are updated; the whole ViT is frozen.}
    \label{fig:architecture}
    \vskip-5pt
\end{figure*}

\mypara{Comparison to closely related work.} Besides INTR~\cite{paul2024simple}, our class-specific attentions are inspired by two other works in different contexts, MCTformer for weakly supervised semantic segmentation \cite{xu2022multi} and Query2Label for multi-label classification \cite{liu2021query2label}. Both of them learned class-specific tokens but aimed to localize visually distinct common objects (\eg, people, horses, and flights). In contrast, we focus on fine-grained analysis: supervised by class labels of visually similar objects (\eg, bird species), we aim to localize their traits (\eg, red spots on wings). One particular feature of \Ours is its \emph{simplicity}, in both implementation and compatibility with pre-trained backbones, without extra modules, loss terms, and changes to the backbones, making it an almost plug-and-pay approach to interpretation. 

Due to space constraints, we provide a detailed related work section in the Supplementary Material (Suppl.).

\section{Approach}
\label{sec:method}

We propose \textbf{Prompt Class Attention Map (\Ours)} to leverage pre-trained Vision Transformers (ViTs)~\cite{dosovitskiy2021an} for fine-grained analysis. The goal is to identify and localize traits that highlight an object category’s identity. \Ours adds learnable class-specific tokens to prompt ViTs, producing class-specific attention maps that reveal traits. 
The overall framework is presented in~\autoref{fig:architecture}.  \emph{We deliberately follow the notation and naming of Visual Prompt Tuning (VPT)~\cite{jia2022visual} for ease of reference.}  

\subsection{Preliminaries}
\label{ss:prep}

A ViT typically contains $N$ Transformer layers~\cite{vaswani2017attention}. Each consists of a Multi-head Self-Attention (MSA) block, a Multi-Layer Perceptron (MLP)
block, and several other operations like layer normalization and residual connections. 

The input image $\mI$ to ViTs is first divided into $M$ fixed-sized patches. Each is then projected into a $D$-dimensional feature space with positional encoding, denoted by $\ve_0^{j}$, with $1\leq j \leq M$. We use $\mE_0=[\ve_0^{1}, \cdots, \ve_0^{M}]\in\R^{D\times M}$ to denote their column-wise concatenation.  

Together with a learnable [CLS] token $\vx_0\in\R^D$, the whole ViT is formulated as:
\begin{align}
[\mE_i, \vx_i] = L_i([\mE_{i-1}, \vx_{i-1} ]), \quad i = 1, \cdots, N, \nonumber
\end{align}
where $L_i$ denotes the $i$-th Transformer layer. The final $\vx_N$ is typically used to represent the whole image and fed into a prediction head for classification. 

\subsection{Prompt Class Attention Map (\Ours)}
\label{ss:P-CAM}

Given a pre-trained ViT and a downstream classification dataset with $C$ classes, we introduce a set of $C$ learnable $D$-dimensional vectors to prompt the ViT. These vectors are learned to be ``class-specific'' by minimizing the cross-entropy loss, during which the ViT backbone is frozen. In the following, we first introduce the baseline version.

\mypara{\OursS.} The $C$  class-specific prompts are injected into the first Transformer layer $L_1$. We denote each prompt by $\vp^{c}\in\R^D$, where $1\leq c\leq C$, and use $\mP = [\vp^{1},\cdots,\vp^{C}]\in\R^{D\times C}$ to indicate their column-wise concatenation. The prompted ViT is:
\begin{align}
[\mZ_1, \mE_1, \vx_1] & = L_1([\mP, \mE_{0}, \vx_{0}]) \nonumber\\
[\mZ_i, \mE_i, \vx_i] & = L_i([\mZ_{i-1},  \mE_{i-1}, \vx_{i-1}]), \quad  i = 2, \cdots, N, \nonumber
\end{align}
where $\mZ_i$ represents the features corresponding to $\mP$, computed by the $i$-th Transformer layer $L_i$. The order among $\vx_{0}$, $\mE_{0}$, and $\mP$ does not matter since the positional encoding of patch locations has already been inserted into $\mE_{0}$. 

To make $\mP = [\vp^1,\cdots,\vp^C]$ class-specific, we employ a cross-entropy loss on top of the corresponding ViT's output, \ie, $\mZ_N = [\vz_N^{1}, \cdots, \vz_N^{C}]$. Given a labeled training example $(\mI, y\in\{1,\cdots, C\})$, we calculate the logit of each class by:
\begin{align}
s[c] = \vw^\top \vz_N^{c}, \quad 1\leq c \leq C, \label{eq:score_rule}
\end{align}
where $\vw\in\R^D$ is a learnable vector. $\mP$ can then be updated by minimizing the loss:
\begin{align}
-\log\left(\cfrac{\exp{\left(s[y]\right)}}{\sum_c \exp{\left(s[c]\right)}}\right). \label{eq:loss}
\end{align}
\mypara{\OursD.} While straightforward, \OursS has two potential drawbacks. First, the class-specific prompts attend to every layer's patch features, \ie, $\mE_i$,  $i = 0, \cdots,  N-1$. However, features of the early layers are often not informative enough but noisy for differentiating classes. Second, the prompts $\vp^1,\cdots,\vp^C$ have a ``double duty.'' Individually, each needs to highlight class-specific traits. Collectively, they need to adapt pre-trained ViTs to downstream tasks, which is the original purpose of VPT~\cite{jia2022visual}. In our case, the downstream task is \emph{a new usage of ViTs on a specific fine-grained dataset.}
  
To address these issues, we resort to the VPT-Deep's design while deliberately \emph{decoupling} injected prompts' roles. VPT-Deep adds learnable prompts to every layer's input. Denote by $\mP_{i-1}=[\vp_{i-1}^1,\cdots,\vp_{i-1}^C]$ the prompts to the $i$-th Transformer layer, the deep-prompted ViT is formulated as:  
\begin{align}
[\mZ_i, \mE_i, \vx_i] & = L_i([\mP_{i-1}, \mE_{i-1}, \vx_{i-1}]), \quad i = 1, \cdots,  N,\label{eq:VPT-Deep}
\end{align}
It is worth noting that the features $\mZ_i$ after the $i$-th layer are not inputted to the next layer, and are typically disregarded. 

In \OursD, we repurpose $\mZ_N$ for classification, following~\autoref{eq:score_rule}. As such, after minimizing the cross entropy loss in~\autoref{eq:loss}, the corresponding prompts $\mP_{N-1}=[\vp_{N-1}^1,\cdots,\vp_{N-1}^C]$ will be \emph{class-specific}. Prompts to the other layers' inputs, \ie, $\mP_{i}=[\vp_{i}^1,\cdots,\vp_{i}^C]$ for $i = 0, \cdots, N-2$, remain \emph{class-agnostic}, because $\vp_{i}^c$ does not particularly serve for the $c$-th class, unlike $\vp_{N-1}^c$. \emph{In other words, \OursD learns both class-specific prompts for trait localization and class-agnostic prompts for adaptation.} The class-specific prompts $\mP_{N-1}$ only attend to the patch features $\mE_{N-1}$ inputted to the last Transformer layer $L_N$, further addressing the other issue in \OursS. 

\emph{In the following, we focus on \OursD.}

\subsection{Trait Identification and Localization}
\label{ss:vis}

During inference, given an image $\mI$, \OursD extracts patch embeddings $\mE_0=[\ve_0^{1}, \cdots, \ve_0^{M}]$ and follows 
\autoref{eq:VPT-Deep} to obtain $\mZ_N$ and \autoref{eq:score_rule} to obtain $s[c]$ for $c\in\{1,\cdots, C\}$. The predicted label $\hat{y}$ is:
\begin{align}
\hat{y} = \argmax_{c\in\{1,\cdots, C\}} s[c].
\end{align}

\mypara{What are the traits of class $c$?} To answer this question, one could collect images whose true and predicted classes are both class $c$ (\ie, correctly classified) and visualize the multi-head attention maps queried by $\vp_{N-1}^c$ in layer $L_N$. 

Specifically, in layer $L_N$ with $R$ attention heads, the patch features $\mE_{N-1}\in\R^{D\times M}$ are projected into $R$ key matrices, denoted by $\mK_{N-1}^r\in\R^{D'\times M}$, $r = 1, \cdots, R$.
The $j$-th column corresponds to the $j$-th patch in $\mI$. Meanwhile, the prompt $\vp_{N-1}^c$ is projected into $R$ query vectors $\vq_{N-1}^{c,r}\in\R^{D'}$, $r = 1, \cdots, R$. Queried by $\vp_{N-1}^c$, the $r$-th head's attention map $\valpha^{c,r}_{N-1}\in\R^M$  is computed by:
\begin{align}
\valpha^{c,r}_{N-1} = \softmax\left(\cfrac{{\mK_{N-1}^r}^\top\vq^{c,r}_{N-1}}{D'}\right)\in\R^M.\label{eq:attention_map}
\end{align}
Conceptually, from the $r$-th head's perspective, the weight $\alpha^{c,r}_{N-1}[j]$ indicates how important the $j$-th patch is for classifying class $c$, hence localizing traits in the image. Ideally, each head should attend to different (sets of) patches to look for multiple traits that together highlight class $c$'s identity. By visualizing each attention map $\valpha^{c,r}_{N-1}$, $r = 1, \cdots, R$, 
instead of pooling them averagely, \Ours can potentially identify up to $R$ different traits for class $c$. 
 
\mypara{Which traits are more discriminative?} For categories that are so distinctive, like ``Red-winged Blackbird,'' a few traits are sufficient to distinguish them from others. To automatically identify these most discriminative traits, we take a greedy approach, \emph{progressively blurring} the least important attention maps until the image is misclassified. The remaining ones highlight traits that are sufficient for classification.

Suppose class $c$ is the true class and the image is correctly classified. In each greedy step, for each of the unblurred heads indexed by $r'$, we iteratively replace $\valpha^{c,r'}_{N-1}$ with $\frac{1}{M}\textbf{1}$ and recalculate $s[c]$ in \autoref{eq:score_rule}, where $\textbf{1}\in\R^M$ is an all-one vector. Doing so essentially blurs the $r'$-th head for class $c$, preventing it from focusing. The head with the \emph{highest blurred $s[c]$} is thus the \emph{least} important, as blurring it degrades classification the least. See Suppl.~for details.

\mypara{Why is an image wrongly classified?}
When $\hat{y}\neq y$ for a labeled image $(\mI,y)$, one could visualize both $\{\valpha^{y,r}_{N-1}\}_{r=1}^R$ and $\{\valpha^{\hat{y},r}_{N-1}\}_{r=1}^R$ to understand why the classifier made such a prediction. For example, some traits of class $y$ may be invisible or unclear in $\mI$; the object in $\mI$ may possess class $\hat{y}$'s visual traits, for example, due to light conditions. 

\subsection{Variants and Extensions}
\label{ss:other}
\mypara{Other \Ours designs.} Besides injecting class-specific prompts to the first layer (\ie, \OursS) or the last (\ie, \OursD), we also explore their interpolation. We introduce class-specific prompts like \OursS to the $i$-th layer and class-agnostic prompts like \OursD to the first $i-1$ layers. See the Suppl.~for a comparison.

\mypara{\Ours for discovering taxonomy keys.} So far, we have focused on a ``flat'' comparison over all the categories. In domains like biology that are full of fine-grained categories, researchers often have built hierarchical decision trees to ease manual categorization, such as taxonomy. The role of each intermediate ``tree node'' is to dichotomize a subset of categories into multiple groups, each possessing certain \emph{group-level} characteristics (\ie, taxonomy keys).       

The \emph{simplicity} of \Ours allows us to efficiently train multiple sets of prompts, one for each intermediate tree node, potentially \emph{(re-)discovering} the taxonomy keys. One just needs to relabel categories of the same group by a single label, before training. In expectation, along the path from the root to a leaf node, each of the intermediate tree nodes should look at different group-level traits on the same image of that leaf node. See~\autoref{fig:hieriarchial_trait} for a preliminary result.
 
\subsection{What is \Ours suited for?}
\label{sec: suitable}
As our paper is titled, \Ours is dedicated to fine-grained \emph{analysis}, aiming to identify and, more importantly, \emph{localize} traits useful for differentiating categories. This, however, does not mean that \Ours would excel in fine-grained classification \emph{accuracy}. Modern neural networks easily have millions if not billions of parameters. How a model predicts is thus still an unanswered question, at least, not fully. It is known if a model is trained mainly to chase accuracies with no constraints, it will inevitably discover ``shortcuts'' in the collected data that are useful for classification but not analysis~\cite{deng2024robust, jackson1991spectre}. 
We thus argue:
\begin{center}
\color{blue}
\emph{
To make a model suitable for fine-grained analysis, one must constrain its capacity, while knowing that doing so would unavoidably hurt its classification accuracy.
}
\end{center}

\Ours is designed with this mindset. Unlike conventional classifiers that employ a fully connected layer on top, \Ours follows~\cite{paul2024simple} and learns a shared vector $\vw$ in~\autoref{eq:score_rule}. The goal of $\vw$ is NOT to capture class-specific information BUT to answer a ``binary'' question: \emph{Based on where a class-specific prompt attends, does the class recognize itself in the input image?}

To elucidate the difference, let us consider a \emph{simplified} single-head-attention Transformer layer with no layer normalization, residual connection, MLP block, and other nonlinear operations. Let $\mV = \{\vv^1, \cdots, \vv^M\}\in\R^{D\times M}$ be the $M$ input patches' value features, $\valpha^c\in\R^M$ be the attention weights of class $c$, and $\valpha^\star\in\R^M$ be the attention weights of the [CLS] token. Conventional models predict classes by:
\begin{align}
\hat{y} = & \argmax_{c} \vw_c^\top (\sum_j \alpha^\star[j] \times \vv^j)\nonumber \\
= & \argmax_{c} \sum_j \alpha^\star[j] \times (\vw_c^\top \vv^j),\label{eq:standard}
\end{align}
where $\vw_c$ stores the fully connected weights for class $c$. We argue that this formulation allows for a potential ``detour,'' enabling the model to correctly classify an image $\mI$ of class $y$ even without meaningful attention weights. In essence, the model can choose to produce holistically discriminative value features from $\mI$ without preserving spatial resolution, such that $\vv^j$ aligns with $\vw_y$ but $\vv^j = \vv^{j'}, \forall j\neq j'$. In this case, regardless of the specific values of  $\valpha^\star$, as long as they sum to one---as is default in the $\softmax$ formulation---the prediction remains unaffected. 

In contrast, \Ours predicts classes by:
\begin{align}
\hat{y} = & \argmax_{c} \vw^\top (\sum_j \alpha^c[j] \times \vv^j)\nonumber\\
= & \argmax_{c} \sum_j \alpha^c[j] \times (\vw^\top \vv^j),\label{eq:INTR}
\end{align}
where $\vw$ is the shared binary classifier. (For brevity, we assume no self-attention among the prompts.) While the difference between \autoref{eq:INTR} and \autoref{eq:standard} is subtle at first glance, it fundamentally changes the model's behavior. In essence, it becomes less effective to store class discriminative information in the channels of $\vv^j$, because there is no $\vw_c$ to align with. Moreover, the model can no longer produce holistic features with no spatial resolution; otherwise, it cannot distinguish among classes since all of their scores $s[c]$ will be exactly the same, no matter what $\valpha^c$ is. 

In response, the model must be equipped with two capabilities to minimize the cross-entropy error:
\begin{itemize}[nosep,topsep=2pt,parsep=0pt,partopsep=2pt, leftmargin=*]
\item Generate localized features $\vv^j$ that highlight discriminative patches (\eg, the red spot on the wing) of an image. 
\item Generate distinctive attention weights $\valpha^c$ across classes, each focusing on traits frequently seen in class $c$.
\end{itemize}
These properties are what fine-grained analysis needs.

In sum, \Ours discourages patch features from encoding class-discriminative holistic information (\eg, the whole object shapes or mysterious long-distance pixel correlations), even if such information can be ``beneficial'' to a conventional classifier. To this end, \Ours needs to \emph{distill} localized, trait-specific information from the pre-trained ViT's patch features, which is achieved through the injected class-agnostic prompts in \OursD.

\begin{figure*}[t]
\centering
\begin{subfigure}[t]{0.45\textwidth}
\includegraphics[width=\textwidth]{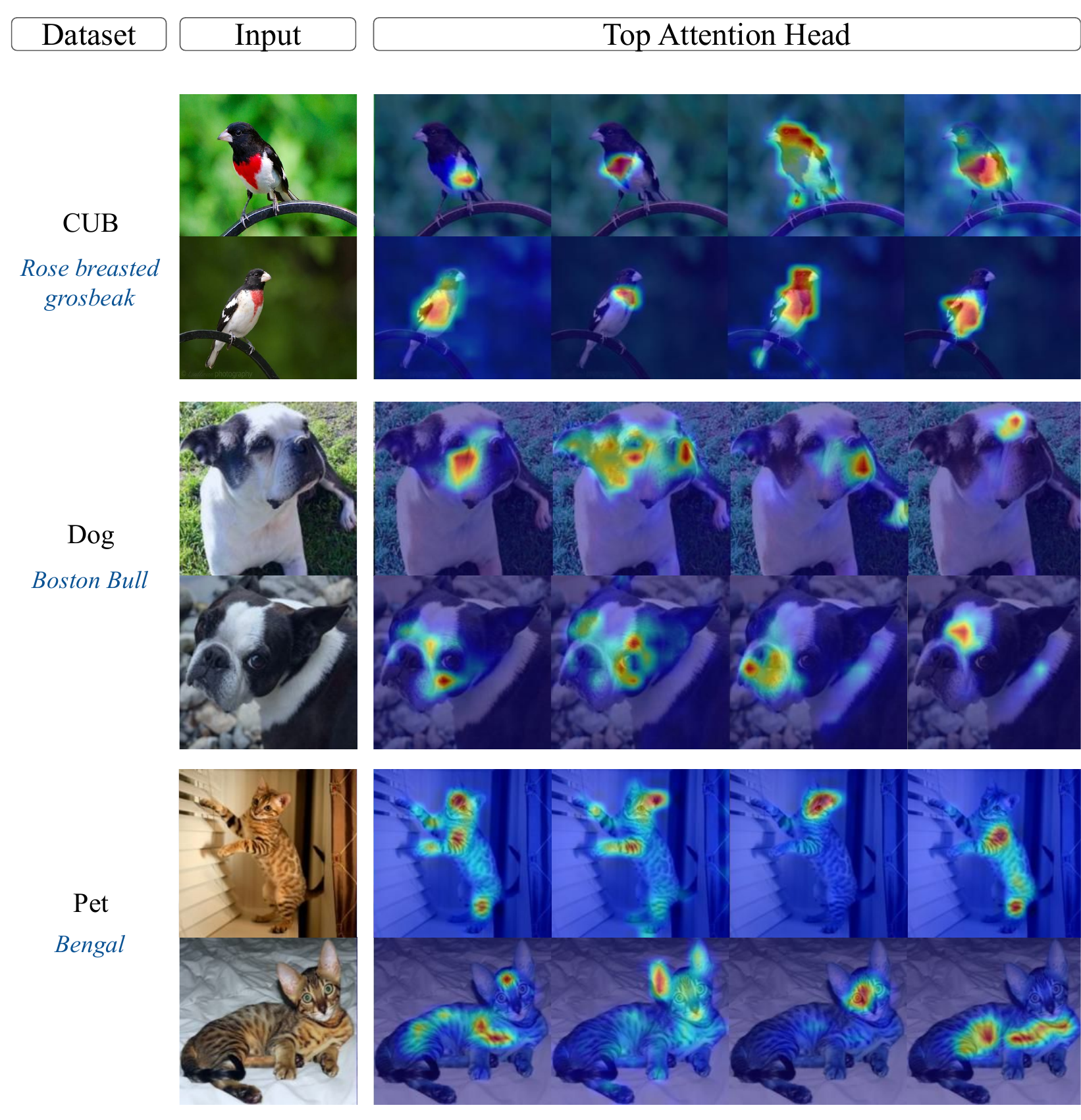}
\end{subfigure}
\begin{subfigure}[t]{0.45\textwidth}
\includegraphics[width=\textwidth]{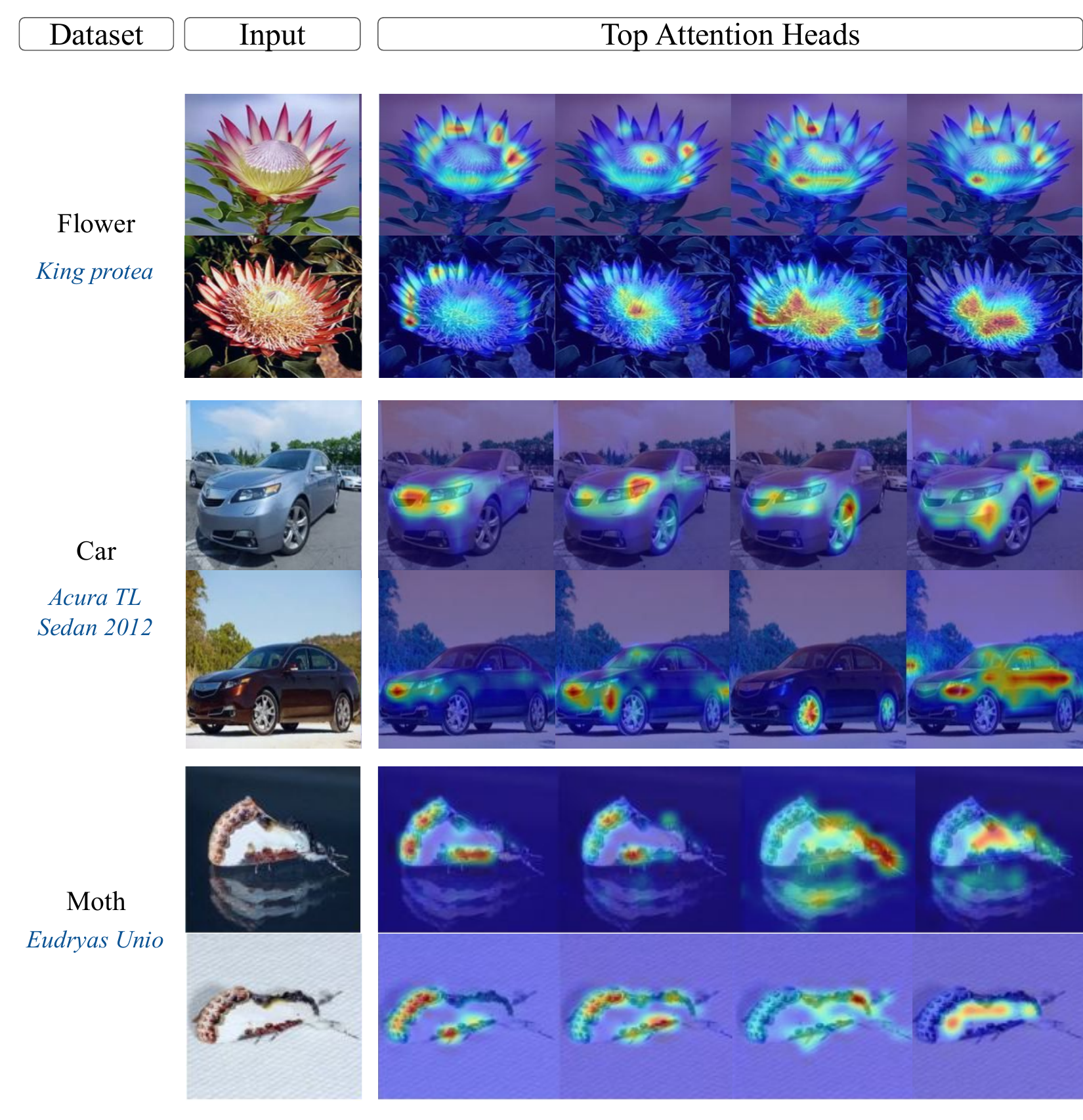}
\end{subfigure}
\vskip-5pt
\caption{\small \textbf{Visualization of \Ours on different datasets.} We show the top four attention maps (from left to right) per correctly classified test example triggered
by the ground-truth classes.}
\vskip-10pt
\label{fig: all_dataset_figure}
\end{figure*}

\section{Experiments}
\label{sec:experiment}

\subsection{Experimental Setup}
\label{sub_sec:experiment_settings}

\textbf{Dataset.}
We comprehensively evaluate the performance of \Ours on \textbf{13} diverse fine-grained image classification datasets across three domains:  \textbf{(1) animal-based}:  CUB-200-2011 (\textit{CUB})~\cite{wah2011caltech}, Birds-525 (\textit{Bird})~\cite{piosenka2023birds},  Stanford Dogs (\textit{Dog})~\cite{khosla2011novel}, Oxford Pet (\textit{Pet})~\cite{parkhi2012cats}, iNaturalist-2021-Moths (\textit{Moth})~\cite{van2021benchmarking}, Fish Vista (\textit{Fish})~\cite{mehrab2024fish}, Rare Species (\textit{RareS.})~\cite{rare_species_dataset} and Insects-2 (\textit{Insects})~\cite{wu2019ip102}; \textbf{(2) plant and fungi-based}: iNaturalist-2021-Fungi (\textit{Fungi})~\cite{van2021benchmarking}, Oxford Flowers (\textit{Flower})~\cite{nilsback2008automated} and Medicinal Leaf (\textit{MedLeaf})~\cite{roopashree2020medicinal}; \textbf{(3) object-based}: Stanford Cars (\textit{Car})~\cite{krause20133d} and Food 101 (\textit{Food})~\cite{bossard2014food}. We provide details about data processing and statistics in  Suppl.

\mypara{Model.}
We consider three pre-trained ViT backbones, DINO~\cite{caron2021emerging}, DINOv2~\cite{oquab2023dinov2}, and BioCLIP~\cite{stevens2024bioclip} across different scales including ViT-B (the main one we use) and ViT-S. 
The backbones are kept completely frozen when applying \Ours. We mainly used DINO, unless stated otherwise. More details can be found in Suppl.

\mypara{Baseline Methods.}
We compared \Ours with explainable methods like  Grad-CAM~\cite{selvaraju2017grad},  Layer-CAM~\cite{jiang2021layercam} and Eigen-CAM~\cite{muhammad2020eigen} as well as with interpretable methods like ProtoPFormer~\cite{xue2022protopformer}, TesNet~\cite{wang2021interpretable}, ProtoConcepts~\cite{ma2024looks} and INTR~\cite{paul2024simple}. More details are in Suppl.

\subsection{Experiment Results}
\label{sub_sec:experiment_results}

\mypara{Is \Ours faithful?}
We first investigate whether \Ours  highlights the image regions that the corresponding classifier focuses on when making predictions.
We use \Ours to rank pixels based on the aggregated attention maps over the top heads. We then employ the insertion and deletion metrics~\cite{petsiuk1806rise}, manipulating highly ranked pixels to measure confidence increase and drop. 

For comparison, we consider post-hoc explainable methods like Grad-CAM~\cite{selvaraju2017grad}, Eigen-CAM~\cite{muhammad2020eigen}, Layer-CAM \cite{jiang2021layercam}, and attention roll-out~\cite{kashefi2023explainability}, based on the same ViT backbone with Linear Probing. 
As summarized in \autoref{tab:insertion_deletion}, \Ours yields higher insertion scores and lower deletion scores, indicating a stronger focus on discriminative image traits and highlighting \Ours's enhanced interpretability over standard post-hoc algorithms.

\begin{table}[t]
\caption{\small Faithfulness evaluation based on insertion and deletion scores. A higher insertion score and a lower deletion score indicate better results. The results are obtained from the validation images
of CUB using the DINO backbone.}
\vskip-5 pt
\centering
\footnotesize
\begin{tabular}{ccc}
\toprule
Method                    & \multicolumn{1}{l}{Insertion$\uparrow$} & \multicolumn{1}{l}{Deletion$\downarrow$} \\
\midrule
Grad-CAM~\cite{selvaraju2017grad}                                 & 0.52                                 & 0.17               \\
Layer-CAM\cite{jiang2021layercam}                                & 0.54                                 & 0.13       \\
Eigen-CAM~\cite{muhammad2020eigen}                                & 0.56                                       & 0.22     
\\      Attention roll-out~\cite{kashefi2023explainability}       & 0.55 & 0.27    
 \\    
\textbf{\Ours} & \textbf{0.61}                                   & \textbf{0.09}  \\
\bottomrule
\end{tabular}
\vskip-5pt

\label{tab:insertion_deletion}
\end{table}

\begin{table}[t]
\footnotesize
\caption{\small {Accuracy (\%) comparison using the DINO backbone.}}
\vskip-8pt
\centering
\begin{tabular}{lcccc}
\toprule
                      & \multicolumn{1}{l}{Bird} & \multicolumn{1}{l}{CUB} & \multicolumn{1}{l}{Dog}      & \multicolumn{1}{l}{Pet}      \\
\midrule                      
Linear Probing           & 99.2                     & 78.6                    & 82.4                         & 92.4                         \\
\Ours & 98.8                     & 73.2	                  & 81.1                          & 91.3  \\           \bottomrule             
\end{tabular}
\vskip-10pt

\label{tab:model_accuracy}
\end{table}

\mypara{\Ours excels in trait identification (human assessment).}
We then conduct a quantitative human study to evaluate trait identification quality for \Ours, TesNet \cite{wang2021interpretable}, and ProtoConcepts \cite{ma2024looks}. 
Participants with no prior knowledge about the algorithms were instructed to compare the expert-identified traits (in text, such as orange belly) and the top heatmaps generated by each method. If an expert-identified trait is seen in the heatmaps, it is considered identified by the algorithm.
On average, participants recognized $60.49\%$ of traits for \Ours, significantly outperforming TesNet and ProtoConcepts whose recognition rates are $39.14\%$ and $30.39$\%, respectively. The results highlight \Ours's superiority in emphasizing and conveying relevant traits effectively. More details are in Suppl.

\begin{figure}[t]
    \centering
    \includegraphics[width=0.9\linewidth]{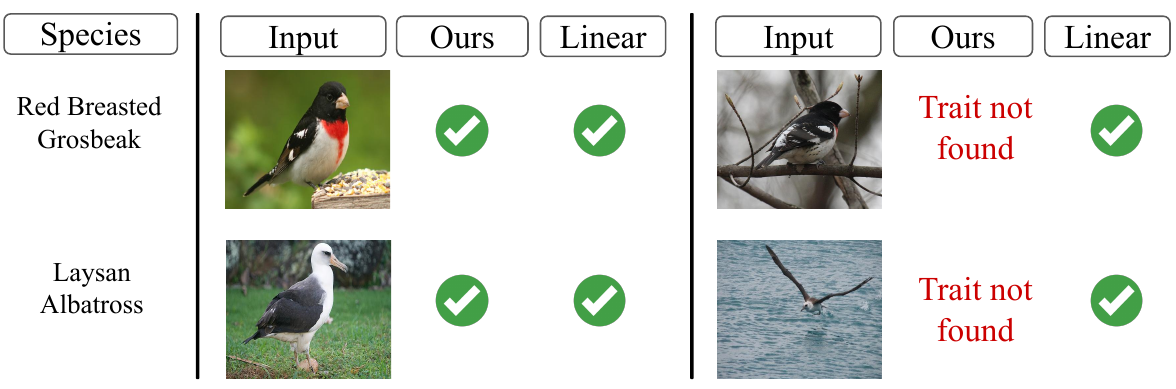}
    \vskip -5pt
    \caption{\small \textbf{Images misclassified by \Ours but correctly classified by Linear Probing.}  Species-specific traits—such as the red breast of  ``Red-breasted Grosbeak"—are barely visible in misclassified images while Linear Probing uses global features such as body shapes, poses, and backgrounds for correct predictions.}
    \label{fig:misclassified_images}
\vskip -8pt
\end{figure}

\mypara{Classification accuracy comparison.}
We observe that \Ours shows a slight accuracy drop compared to Linear Probing (see \autoref{tab:model_accuracy}). However, the images misclassified by \Ours but correctly classified by Linear Probing align with our design philosophy: \Ours classifies images based on the presence of class-specific, localized traits and would fail if they are invisible. 
As shown in \autoref{fig:misclassified_images}, discriminative traits—such as the red breast of the Red-breasted Grosbeak—are barely visible in images misclassified by \Ours due to occlusion, unusual poses, or lighting conditions. Linear Probing correctly classifies them by leveraging global information such as body shapes and backgrounds. Please see more analysis in Suppl.

\begin{figure}[t!]
    \centering
        \includegraphics[width=1\linewidth]{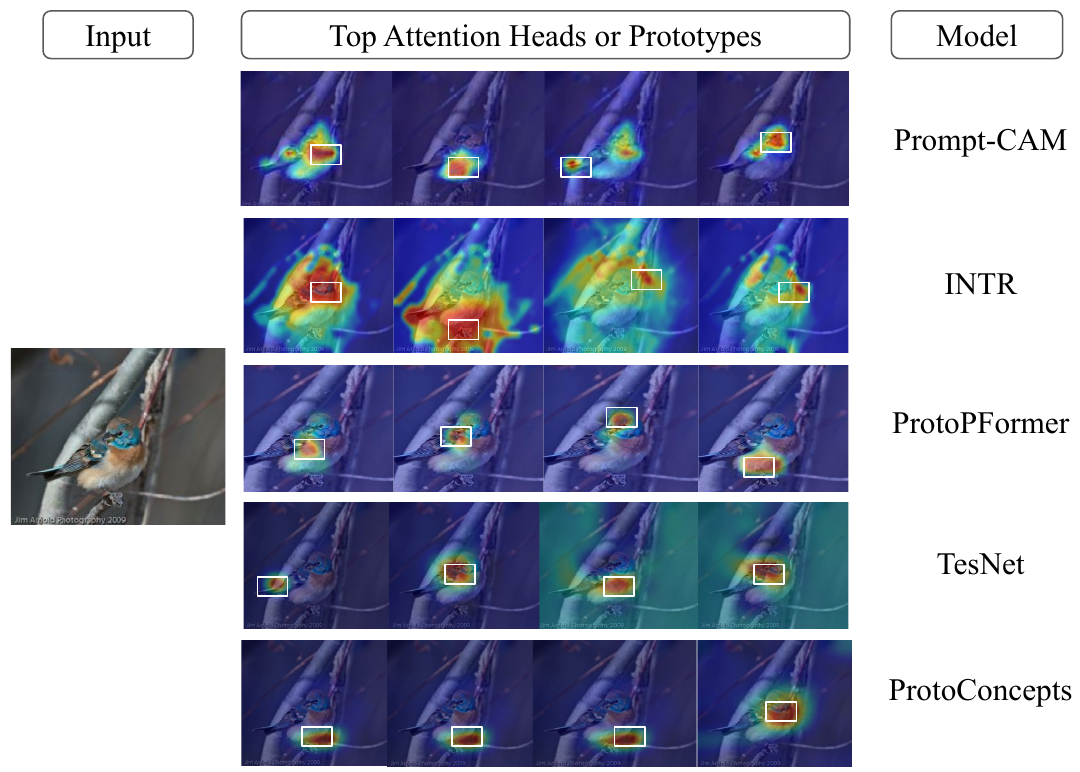}
        \vskip -5pt
    \caption{\small \textbf{Comparison of interpretable models.} Visual demonstration (heatmaps and bounding boxes) of the four most activated responses of attention heads (\Ours and INTR) or prototypes of each method on a ``Lazuli Bunting" example image.}
    \label{fig:interpretableComparison}
\vskip -10pt
\end{figure}

\mypara{Comparison to interpretable models.}
We conduct a qualitative analysis to compare \Ours with other interpretable methods—ProtoPFormer, INTR, TesNet, and ProtoConcepts. \autoref{fig:interpretableComparison} shows the top-ranked attention maps or prototypes generated by each method.
\Ours can capture a more extensive range of distinct, fine-grained traits, in contrast to other methods that often focus on a narrower or repetitive set of attributes (for example, ProtoConcepts in the first three ranks of the fifth row). This highlights \Ours's ability to identify 
and localize different traits that collectively define a category's identity.

\subsection{Further Analysis and Discussion}
\label{sub_sec:experiment_3}

\begin{figure}[t]
    \centering
    \includegraphics[width=.6\linewidth]{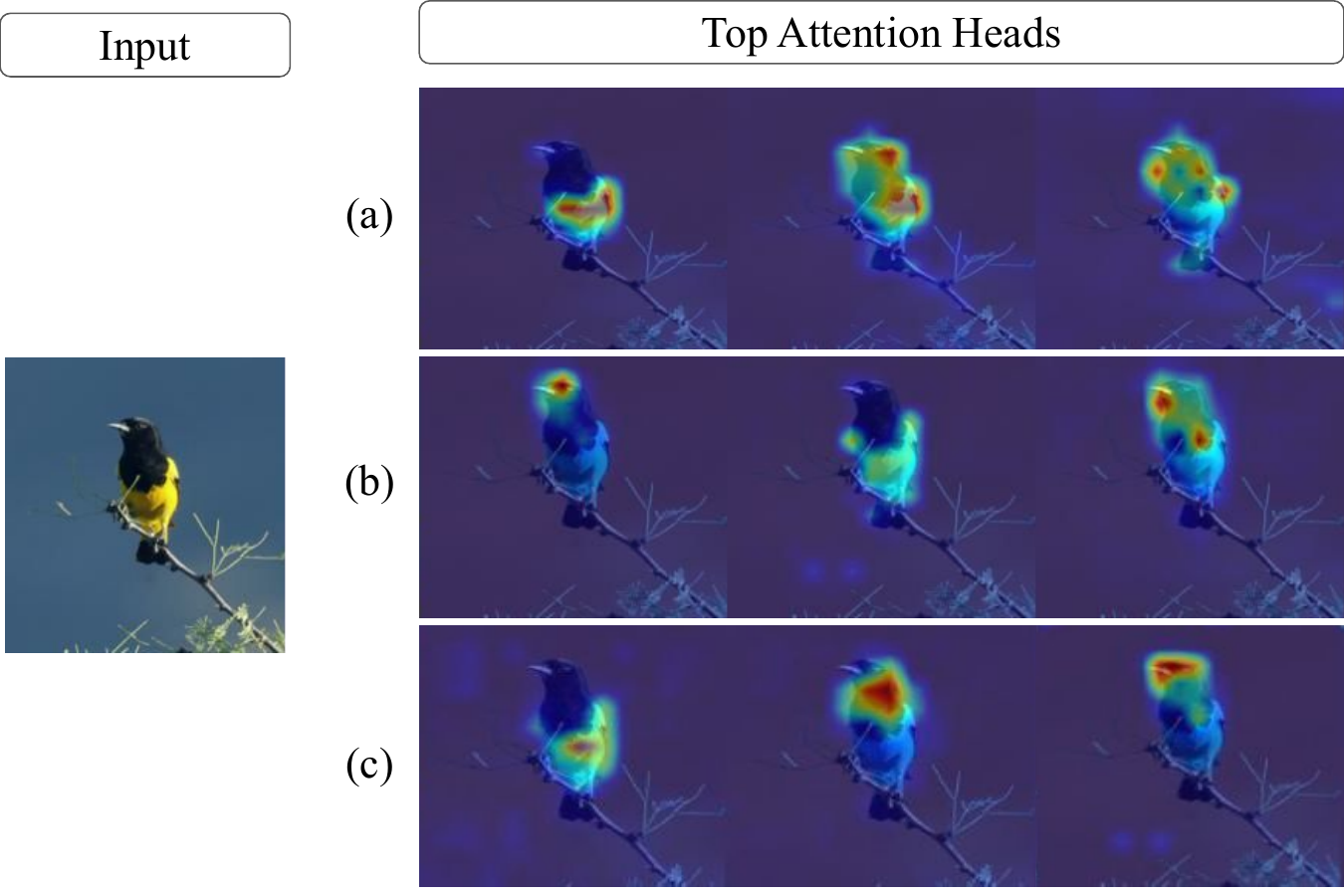}
    \vskip-5pt
\caption{\textbf{\Ours on different backbones}. Here we show the top attention maps for \Ours on (a) DINO, (b) DINOv2, and (c) BioCLIP backbone. All three sets of attention heads point to consistent key traits of the species ``Scott Oriole"---yellow belly, black head, and black chest.}
    \label{fig:dino_vs_dinov2_bioclip}
    \vskip -8pt
\end{figure}

\mypara{\Ours on different backbones.}
\autoref{fig:dino_vs_dinov2_bioclip} illustrates that \Ours is compatible with different ViT backbones. We show the top three attention maps generated by \Ours using different ViT backbones on an image of  Scott Oriole, highlighting consistent identification of traits for species recognition, irrespective of the backbones. Please see the caption and Suppl.~for details.

\mypara{\Ours on different datasets.}
\autoref{fig: all_dataset_figure} presents the top four attention maps generated by \Ours across various datasets spanning diverse domains, including \textit{animals}, \textit{plants}, and \textit{objects}. \Ours effectively captures the most important traits in each case to accurately identify species, demonstrating its remarkable generalizability and wide applicability.

\mypara{\Ours can detect biologically meaningful traits.}
As shown in \autoref{fig: all_dataset_figure}, \Ours consistently identifies traits from images of the same species (\eg, the red breast and white belly for Rose-breasted Grosbeak). This is further demonstrated in  \autoref{fig:teaser} (d), where we progressively mask attention heads (detailed in \autoref{ss:vis}) until the model can no longer generate high-confidence predictions for correctly classifying images of Scott Oriole. The remaining heads 1 and 11 highlight the essential traits, \ie, the black head and yellow belly. \Ours also enables identifying common traits between species. This is achieved by visualizing the image of one class (\eg, Scott Oriole) using other classes' prompts (\eg, Brewer Blackbird or Baltimore Oriole). As shown in \autoref{fig:teaser} (c), Brewer Blackbird shares the head and neck color with Scott Oriole.
These results demonstrate \Ours's ability to recognize species in a biologically meaningful way.

\begin{figure}[t]
    \centering
    \includegraphics[width=1\linewidth]{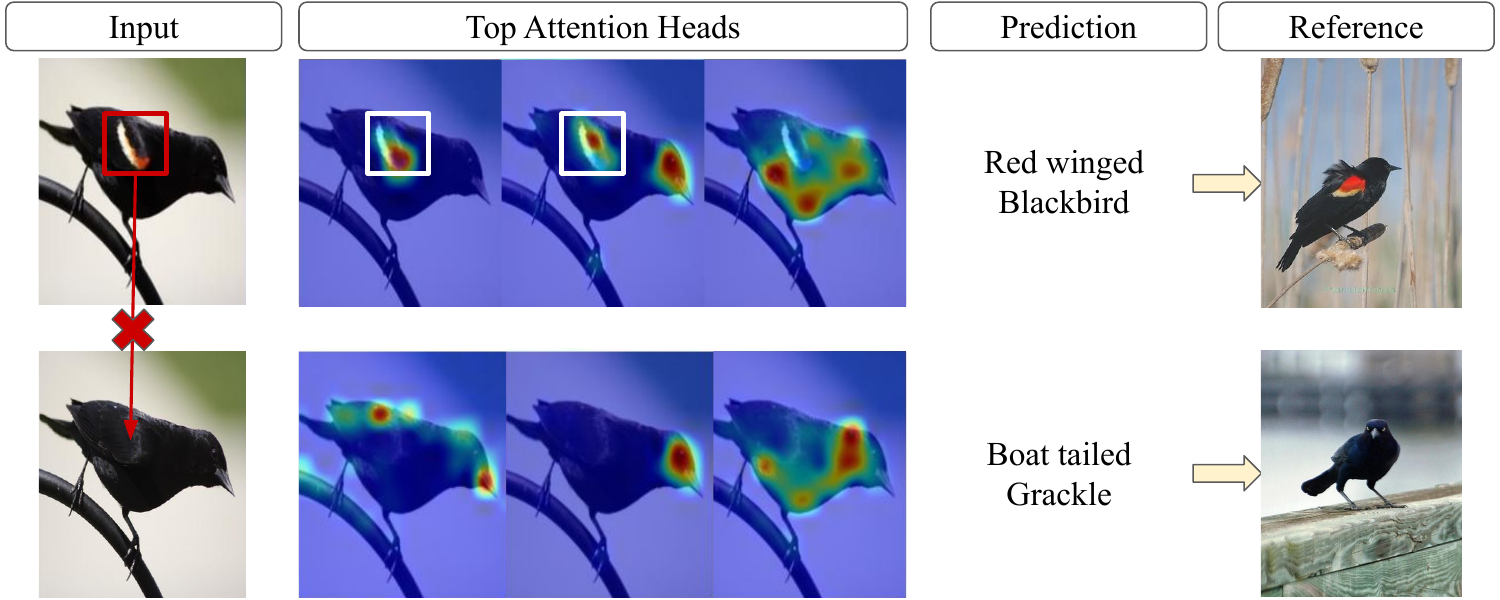}
    \vskip -5pt
    \caption{\textbf{Trait manipulation.} The top row shows attention maps for a correctly classified ``Red-winged Blackbird" image. In the second row, the red spot on the bird's wings was removed, and \Ours subsequently classified it as a ``Boat-tailed Grackle," as depicted in the reference column. }
    \label{fig:trait_manipulation}
    \vskip -10pt
\end{figure}

\mypara{\Ours can identify and interpret trait manipulation.}

We conduct a counterfactual-style analysis to investigate whether \Ours truly relies on the identified traits for making predictions. 
For instance, to correctly classify the Red-winged Blackbird, it highlights the red-wing patch (the first row of \autoref{fig:trait_manipulation}), consistent with the field guide provided by the Cornell Lab of Ornithology. When we remove this red spot from the image to resemble a Boat-tailed Grackle, the model no longer highlights the original position of the red patch. As such, it does not predict the image as a Red-winged Blackbird but a Boat-tailed Grackle (the second row of \autoref{fig:trait_manipulation}). This shows \Ours's sensitivity to trait differences, showcasing its interpretability in fine-grained recognition.

\begin{figure}[t]
    \centering
    \includegraphics[width=1\linewidth]{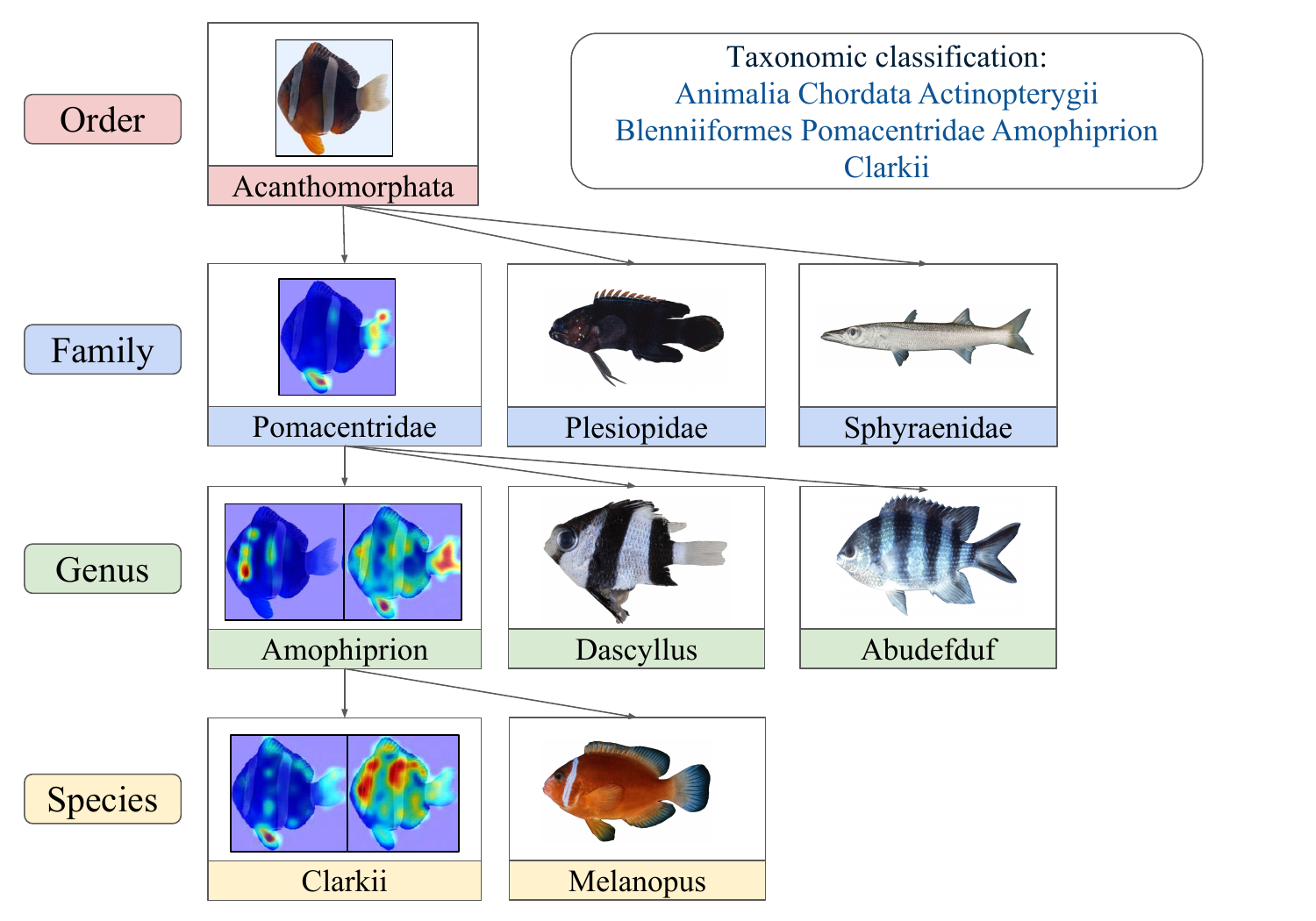}
    \vskip-5pt
    \caption{ \textbf{\Ours can detect taxonomically meaningful traits.} Give an image of the species ``Amophiprion Clarkii,'' \Ours highlights the pelvic fin and double stripe to distinguish it from  ``Amophiprion Melanopus'' at the species level. When it goes to the genus level, \Ours looks at the pattern in the body and tail to classify the image as the ``Amophiprion'' genus. As we go up, fishes at the family level become visually dissimilar. \Ours only needs to look at the tail and pelvic fin to classify the image as the ``Pomacentridae'' family.}
\label{fig:hieriarchial_trait}
\vskip -10pt
\end{figure}
\mypara{\Ours can detect taxonomically meaningful traits.}
We train \Ours based on a hierarchical framework, considering four levels of taxonomic hierarchy: \textit{Order}  
 $\rightarrow$\textit{ Family} 
$\rightarrow$ \textit{Genus} $\rightarrow$ \textit{Species} of Fish Dataset. In this setup, \Ours progressively shifts its focus from coarse-grained traits at the \textit{Family} level to fine-grained traits at the \textit{Species} level to distinguish categories (shown in \autoref{fig:hieriarchial_trait}).
This progression suggests  \Ours's potential to automatically identify and localize taxonomy keys to aid in biological and ecological research domains. We provide more details in Suppl.

\section{Conclusion}
\label{sec:conclusion}

We present Prompt Class Attention Map (\Ours), a simple yet effective interpretable approach that leverages pre-trained ViTs to identify and localize discriminative traits for fine-grained classification. \Ours is easy to implement and train. Extensive empirical studies highlight both the strong performance of \Ours and the promise of repurposing standard models for interpretability.

\clearpage
\section*{Acknowledgment}
This research is supported in part by grants from the National Science Foundation (OAC-2118240, HDR Institute:Imageomics). The authors are grateful for the generous support of the computational resources from the Ohio Supercomputer Center.

{
    \small
    \bibliographystyle{ieeenat_fullname}
    \bibliography{main}
}
\clearpage
\setcounter{page}{1}
\maketitlesupplementary

\appendix
The supplementary is organized as follows.
\begin{itemize}
  \item \autoref{supp:related_work}: Related Work
  \item \autoref{supp:architecture}: Details of Architecture Variant (cf. \autoref{ss:other} of the main paper)
  \item \autoref{supp:dataset}: Dataset Details (cf. \autoref{sub_sec:experiment_settings} of the main paper)
  \item \autoref{supp:inner_work_visualization}: Inner Workings of Visualization (cf. \autoref{ss:vis} of the main paper)
  \item \autoref{supp:exp_settings}: Additional Experiment Settings (cf. \autoref{sub_sec:experiment_settings} of the main paper)
  \item \autoref{supp:exp_results}: Additional Experiment Results and Analysis (cf. \autoref{sub_sec:experiment_results} of the main paper)
  \item \autoref{supp:more_visual}: More visualizations of different dataset (cf. \autoref{fig: all_dataset_figure} of the main paper)
\end{itemize}

\section{Related Work}
\label{supp:related_work}
\mypara{Pre-trained Vision Transformer.}
Vision Transformers (ViT)~\cite{dosovitskiy2021an}, pre-trained on massive amounts of data, has become indispensable to modern AI development. For example, ViTs pre-trained with millions of image-text pairs via a contrastive objective function (\eg, a CLIP-ViT model) show an unprecedented zero-shot capability, robustness to distribution shifts and serve as the encoders for various power generative models (\eg Stable Diffusion~\cite{rombach2022high} and LLaVA~\cite{liu2024visual}). Domain-specific CLIP-based models like BioCLIP~\cite{stevens2024bioclip} and RemoteCLIP~\cite{liu2024remoteclip}, trained on millions of specialized image-text pairs, outperform general-purpose CLIP models within their respective domains. Moreover, ViTs trained with self-supervised objectives on extensive sets of well-curated images, such as DINO and DINOv2~\citep{caron2021emerging,oquab2023dinov2}, effectively capture fine-grained localization features that explicitly reveal object and part boundaries. We employ DINO, DINOv2, and BioCLIP as our backbone models in light of our focus on fine-grained analysis.

\mypara{Prompting Vision Transformer.}
Traditional approaches to adapt pre-trained transformers—full fine-tuning and linear probing—face challenges: the former is computationally intensive and prone to overfitting, while the latter struggles with task-specific adaptation~\cite{maifine,mai2024lessons}. Prompting, first popularized in natural language processing (NLP), addressed such challenges by prepending task-specific instructions to input text, enabling large language models like GPT-3 to perform zero-shot and few-shot learning effectively~\cite{10.5555/3495724.3495883}.

Recently, prompting has been introduced in vision transformers (ViTs) to enable efficient adaptation while leveraging the vast capabilities of pre-trained ViTs~\cite{zhou2022learning, jia2022visual, tu2023visual}. Visual Prompt Tuning (VPT) \citep{jia2022visual} introduces learnable embedding vectors, either in the first transformer layer or across layers, which serve as ``prompts" while keeping the backbone frozen. This offers a lightweight and scalable alternative to full fine-tuning, achieving competitive performance on a diverse range of tasks while preserving the pre-trained features.

\mypara{Explainable methods.}
Understanding the decision-making process of neural networks has gained significant traction, particularly in tasks where model transparency is critical. Explainable methods (XAI) focus on post-hoc analysis to provide insights into pre-trained models without altering their structure. Methods like Class Activation Mapping (CAM)~\cite{zhou2016learning} and Gradient-weighted CAM (Grad-CAM) \citep{selvaraju2017grad} visualize class-specific contributions by projecting gradients onto feature maps. Subsequent improvements, such as Score-CAM \citep{wang2020score} and Eigen-CAM \citep{muhammad2020eigen}, incorporate global feature contributions or principal component analysis to generate more detailed explanations. Despite these advancements, many XAI methods produce coarse, low-resolution heatmaps, which can be imprecise and fail to fully capture the model’s decision-making process.

\mypara{Interpretable methods.}
In contrast, interpretable methods provide a direct understanding of predictions by aligning intermediate representations with human-interpretable concepts. Early approaches such as ProtoPNet \citep{chen2019looks} utilized ``learnable prototypes" to represent class-specific features, enabling visual comparison between input features and prototypical examples. Extensions like ProtoConcepts \citep{ma2024looks}, ProtoPFormer \citep{xue2022protopformer}, and TesNet \citep{wang2021interpretable} have refined this approach, integrating prototypes into transformer-based architectures to achieve higher accuracy and interoperability. More recent advancements leverage transformer architectures to enable interpretable decision-making. For example, Concept Transformers utilize query-based encoder-decoder designs to discover meaningful concepts~\cite{rigotti2021attention}, while methods like INTR \citep{paul2024simple} employ competing query mechanisms to elucidate how the model arrives at specific predictions. While these approaches offer fine-grained interpretability, they require substantial modifications to the backbone, leading to increased training complexity and longer computational times for new datasets.
 
 \Ours aims to overcome the shortcomings of both approaches. The special prediction mechanism encourages explainable, class-specific attention that is aligned well with model predictions. Simultaneously, we leverage pre-trained ViTs by simply modifying the usage of task-specific prompts without altering the backbone architecture.

\section{Details of Architecture Variant}
\label{supp:architecture}
\begin{figure}[t]
    \centering
    \includegraphics[width=1\linewidth]{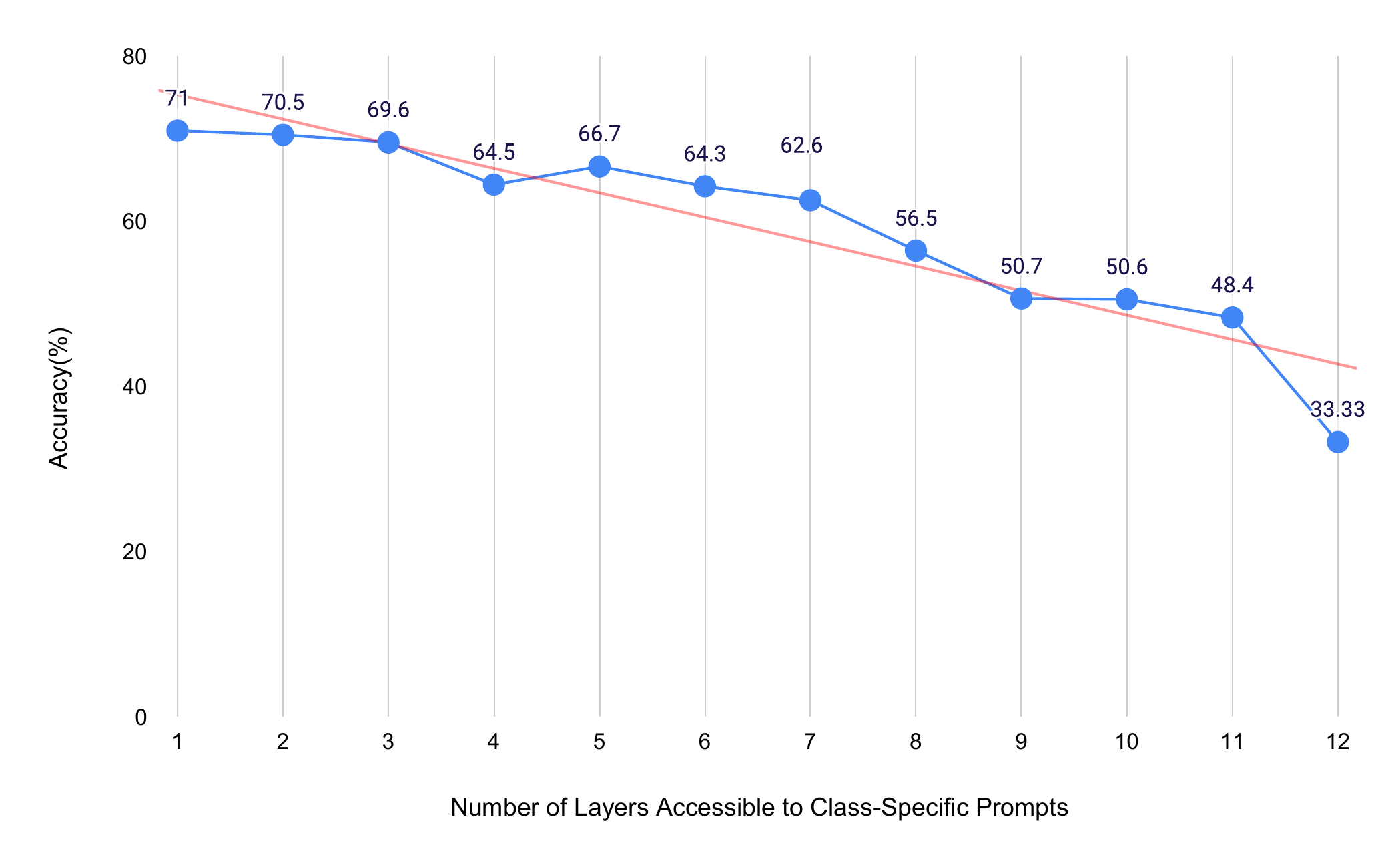}
    \caption{\textbf{Accuracy versus the number of layers (from last layer to first) attended by class-specific prompts.} As the number of attended layers increases in class-specific prompts, accuracy decreases, highlighting the importance of class-agnostic prompts. The more class-agnostic prompts a model has, the better trait localization and higher accuracy are achieved.}
    \label{supp_fig:shallow_to_deep_plot}
    \vskip -8pt
\end{figure}

In this section, we explore variations of \Ours by experimenting with the placement of class-specific prompts within the vision transformer (ViT) architecture. While \OursS introduces class-specific prompts in the first layer and \OursD applies them in the final layer, we also investigate injecting these prompts at various intermediate layers. Specifically, we control the layer depth at which class-specific prompts are added and analyze their impact on feature interpolation.

In \OursS, class-specific prompts are introduced at the first layer ($i=1$), allowing them to interact with patch features across all transformer layers (\ie, $\mE_i$,  $i = 0, \cdots,  N-1$) without using class-agnostic prompts. As we increase the layer index $i$ where class-specific prompts are added, the number of layers class-specific prompts interact decreases. At the same time, the number of preceding class-agnostic prompts increases, which interacts with the preceding $(i-1)$ layers (mentioned in \autoref{ss:P-CAM}).

\begin{figure*}[t]
    \centering
    \includegraphics[width=\linewidth]{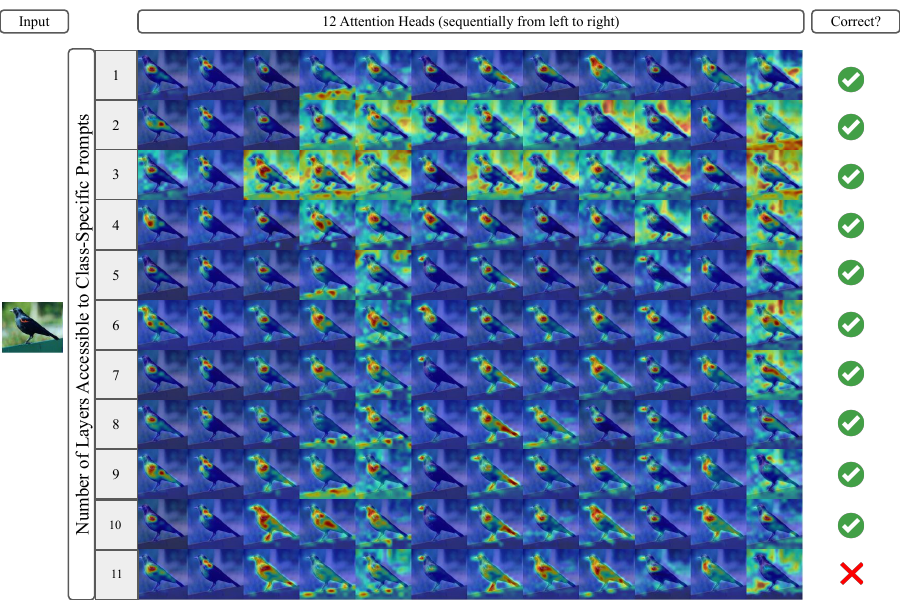}
    \caption{\textbf{Visualization of attention maps for different configurations of \Ours.} For a random image of the ``Red-Winged Blackbird" species, twelve attention heads of the last layer of \Ours on the DINO backbone are shown for the ground truth class prompt. The first row shows class-specific prompts attending to only the last layer (as \OursD), resulting in highly localized attention on fine-grained traits, such as the red patch on the wings of the ``Red-Winged Blackbird." As these prompts attend to increasingly more layers (progressing down the rows), the attention becomes more diffuse, covering broader regions of the object and eventually leading to a loss of focus on relevant traits.}
    \label{supp_fig:shallow_to_deep}
    \vskip -8pt
\end{figure*}

In \autoref{supp_fig:shallow_to_deep}, we demonstrate the relationship between the number of layers accessible to class-specific prompts and their ability to localize fine-grained traits effectively. The visualization provides a clear pattern: as the prompts attend only to the last layer (first row) (same as \OursD), their focus is highly localized on discriminative traits, such as the red patch on the wings of the ``Red-Winged Blackbird." This precise focus enables the model to excel in fine-grained trait analysis.

As we move downward through the rows, class-specific prompts attending to increasingly more layers (from top to bottom), the attention maps become progressively more diffused. For instance, in the middle rows (e.g., rows 6–8), the attention begins to cover broader regions of the object rather than the trait of interest. This diffusion correlates with a drop in accuracy, as seen in the accuracy plot, \autoref{supp_fig:shallow_to_deep_plot}.

In the bottom rows (e.g., rows 10–11), the attention becomes scattered and unfocused, covering irrelevant regions. This fails to correctly classify the object. The accuracy plot confirms this trend: as the class-specific prompts attend to more layers, accuracy steadily decreases.

\section{Dataset Details}
\label{supp:dataset}

\begin{table}[htb]
\centering
\caption{Dataset statistics (Animals).}
\resizebox{\linewidth}{!}{
    \begin{tabular}{l c c c c c c c c}
    \toprule
    \multicolumn{1}{c}{} & \multicolumn{7}{c}{Animals} & \multicolumn{1}{c}{} \\
    \cmidrule(lr){2-8}
    \multicolumn{1}{c}{} & Bird & CUB  & Dog & Pet & Insects & Fish & Moth & RareS. \\
    \midrule
    \# Train Images      & 84,635     & 5,994 & 12,000         & 3,680       & 52,603   & 35,328 & 5,000  & 9,584\\
    \# Test Images       & 2,625      & 5,795 & 8,580          & 3,669       & 22,619    & 7,556 & 1,000 & 2,399\\
    \# Labels            & 525       & 200  & 120           & 37         & 102    & 414 & 100 & 400  \\
    \bottomrule
    \end{tabular}
}

\label{tab:dataset_stats_animals}
\end{table}

\begin{table}[htb]
\centering
\caption{Dataset statistics (Plants \& Fungi and Objects).}
\resizebox{\linewidth}{!}{
    \begin{tabular}{l c c c c c c}
    \toprule
    \multicolumn{1}{c}{} & \multicolumn{3}{c}{Plants \& Fungi} & \multicolumn{2}{c}{Objects} \\
    \cmidrule(lr){2-4} \cmidrule(lr){5-6}
    \multicolumn{1}{c}{} & Flower & MedLeaf & Fungi & Car & Food \\
    \midrule
    \# Train Images      & 2,040           & 1,455   & 12,250        & 8,144            & 75,750     \\
    \# Test Images       & 6,149           & 380        & 2,450    & 8,041            & 25,250     \\
    \# Labels            & 102            & 30       & 245      & 196             & 101       \\
    \bottomrule
    \end{tabular}
}
\label{tab:dataset_stats_plants_objects}
\end{table}

We comprehensively evaluate the performance of \Ours on a diverse set of benchmark datasets curated for fine-grained image classification across multiple domains. The evaluation includes animal-based datasets such as CUB-200-2011 (\textbf{CUB})~\cite{wah2011caltech}, Birds-525 (\textbf{Bird})~\cite{piosenka2023birds},  Stanford Dogs (\textbf{Dog})~\cite{khosla2011novel}, Oxford Pet (\textbf{Pet})~\cite{parkhi2012cats}, iNaturalist-2021-Moths (\textbf{Moth})~\cite{van2021benchmarking}, Fish Vista (\textbf{Fish})~\cite{mehrab2024fish}, Rare Species (\textbf{RareS.})~\cite{rare_species_dataset} and Insects-2 (\textbf{Insects})~\cite{wu2019ip102}. Additionally, we assess performance on plant and fungi-based datasets, including iNaturalist-2021-Fungi (\textbf{Fungi})~\cite{van2021benchmarking}, Oxford Flowers (\textbf{Flower})~\cite{nilsback2008automated} and Medicinal Leaf (\textbf{MedLeaf})~\cite{roopashree2020medicinal}. Finally, object-based datasets, such as Stanford Cars (\textbf{Car})~\cite{krause20133d} and Food 101 (\textbf{Food})~\cite{bossard2014food}, are also included to ensure comprehensive coverage across various fine-grained classification tasks. For the Moth and Fungi dataset, we extract species belonging to Noctuidae Family from taxonomic class \textit{Animalia Arthropoda Insecta Lepidoptera Noctuidae} and species belonging to Agaricomycetes Class from taxonomic path \textit{Fungi$\rightarrow$ Basidiomycota}, respectively, from the iNaturalist-2021 dataset. For hierarchical classification and trait localization, we use taxonomical information from the Fish and iNaturalist-2021 dataset. We provide dataset statistics in \autoref{tab:dataset_stats_animals} and \autoref{tab:dataset_stats_plants_objects}.

\section{Inner Workings of Visualization}
\label{supp:inner_work_visualization}

\begin{figure}[t]
    \centering
    \includegraphics[width=1\linewidth]{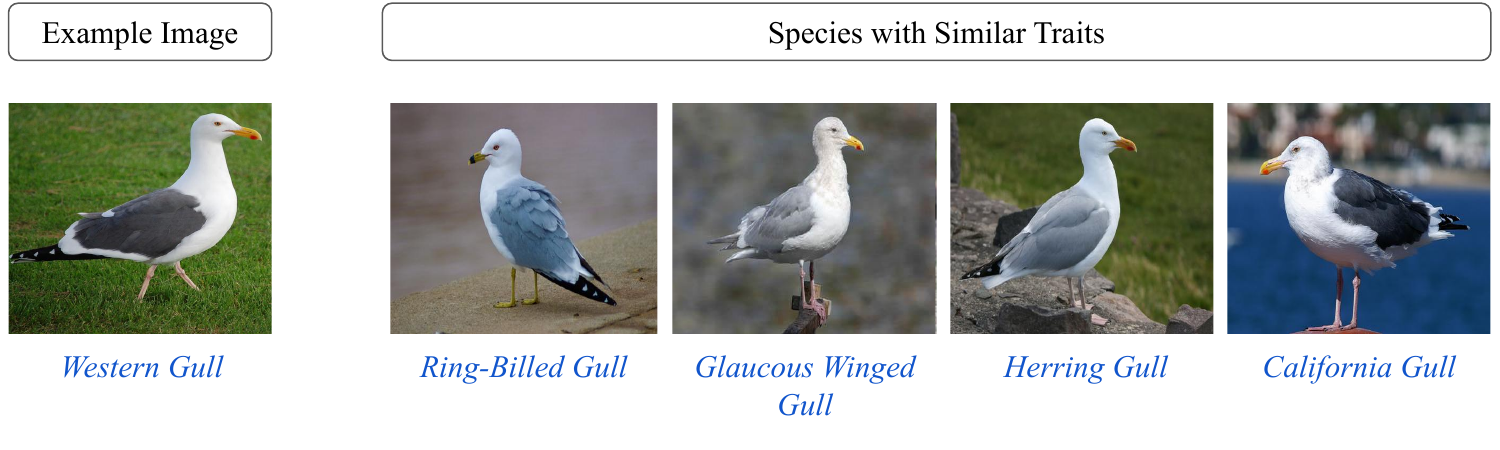}
    \caption{\textbf{Example Image of a ``Western Gull'' and its closest bird species, highlighting overlapping traits.} Correctly classifying the ``Western Gull'' requires attention to multiple subtle traits, as it shares many traits with similar species. This highlights the need to examine a broader range of attributes for accurate classification. }
    \label{supp_fig:close_birds_of_a_species}
    \vskip -8pt
\end{figure}
\mypara{Which traits are more discriminative?}
\label{sup_ss:traits_discriminative}
As discussed in \autoref{ss:vis}, certain categories within the CUB dataset exhibit distinctive traits that are highly discriminative. For instance, in the case of the ``Red-winged Blackbird," the defining features are its red-spotted black wings. Similarly, the ``Ruby-throated Hummingbird" is characterized by its ruby-colored throat and sharp, long beak. However, some species require consideration of multiple traits to distinguish them from others. For example, correctly classifying a ``Western Gull'' demands attention to several subtle traits (\autoref{supp_fig:close_birds_of_a_species}), as it shares many features with other species. This observation raises a key question: can we automatically identify and rank the most important traits for a given image of a species?

\begin{figure*}[t]
    \centering
    \includegraphics[width=.8\linewidth]{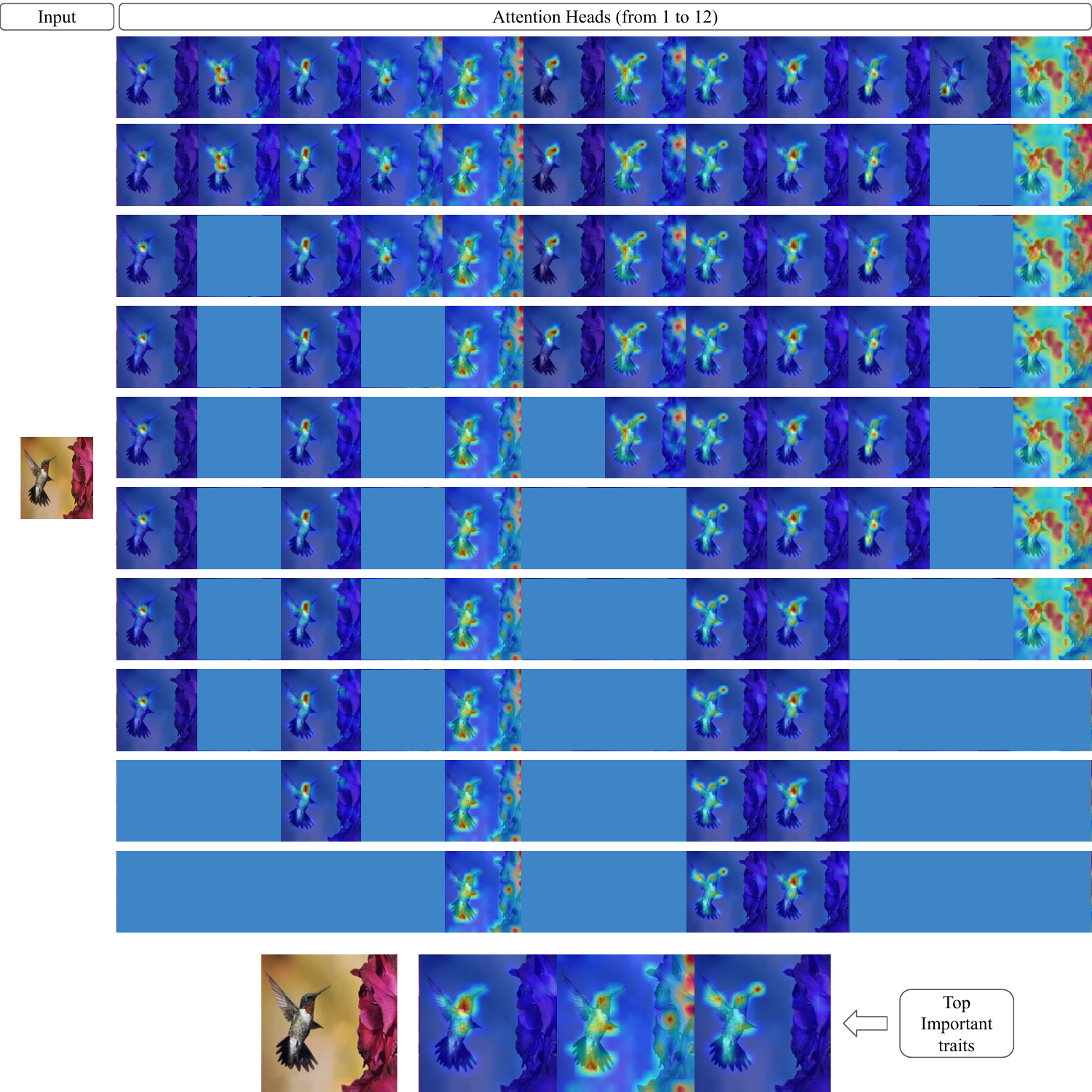}
    \caption{\textbf{Greedy approach to identify and rank important traits for species classification.} For the species ``Ruby Throated Hummingbird", we progressively blur attention heads (from top to bottom), retaining only the traits necessary for correct classification, using the \Ours on the DINO backbone. The blurred attention heads are shown in solid blue color.}
    \label{supp_fig:greedy_approach}
    \vskip -8pt
\end{figure*}

To address this, we propose a greedy algorithm that progressively ``blurs" traits in a correctly classified image until its decision changes. This process reveals the traits that are both necessary and sufficient for the correct prediction.

\begin{figure*}[t]
    \centering
    \includegraphics[width=0.8\linewidth]{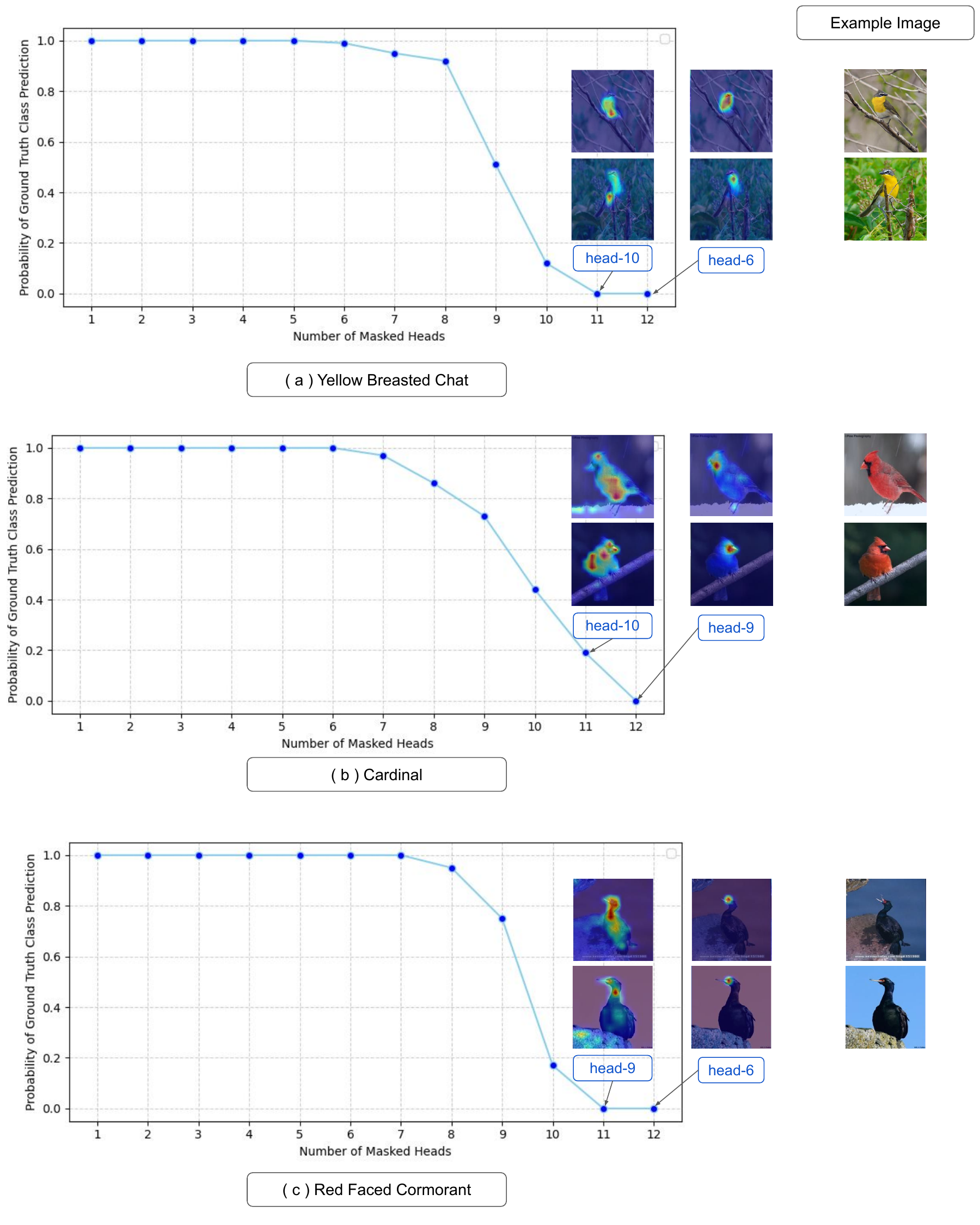}
    \caption{\textbf{Visualization of ground truth class probability vs. the number of masked heads at the species level in \Ours.} The left plots show how the probability of the ground truth class changes for all correctly classified images in a species, as heads are progressively masked in the greedy approach discussed in \autoref{supp:inner_work_visualization}. For class (a) ``Yellow Breasted Chat,'' the probability drops significantly after masking eight heads, indicating that the last four heads are critical. The top two heads, head-6 and head-10, focus on the yellow breast and lower belly. For class (b) ``Cardinal,'' the top 2 heads, head-9 and head-10, attend to the black pattern on the face and the red belly. In class (c) ``Red Faced Cormorant,'' the critical heads, head-6 and head-9, emphasize the red head and the neck's shape. These results highlight the interpretability of \Ours in identifying essential traits for each species.}
    \label{supp_fig:head_vs_probablity}
    \vskip -8pt
\end{figure*}

\mypara{Greedy approach for identifying discriminative traits:} Suppose class $c$ is the true class and the image is correctly classified. In the first greedy step, for each attention head, $r = 1, \cdots, R$ (R attention heads), we iteratively replace the attention vector $\valpha^{c,r}_{N-1}$, with a uniform distribution:
\[
\valpha^{c,r}_{N-1} \leftarrow \frac{1}{M} \mathbf{1},
\]
where $\mathbf{1} \in \mathbb{R}^M $ is a vector of all ones, and $M$ is the number of patches. This replacement effectively assigns equal importance to all patches in the attention weights, thereby ``blurring" the $r$-th head’s contribution to class $c$. After this modification, we recalculate the score $s[c]$ in \autoref{eq:score_rule}.

For each iteration, we select the attention head \( r^* \) that, when blurred, results in the highest probability for the correct class \( c \). This head \( r^* \) is then added to \( B_a \) (set of blurred attention heads), as the \emph{blurred} head with the \emph{highest $s[c]$}  is the \emph{least} important and contributes the least discriminative information for class \( c \). We repeat this process, iteratively blurring additional heads and updating \( B_a \), until blurring any remaining head not in \( B_a \) changes the model’s prediction.
 In \autoref{supp_fig:greedy_approach}, for an image of ``Ruby Throated Hummingbird" we show this greedy approach, by progressively blurring out the attention heads in each step, retaining only necessary traits.

\begin{figure}[t]
    \centering
    \includegraphics[width=\linewidth]{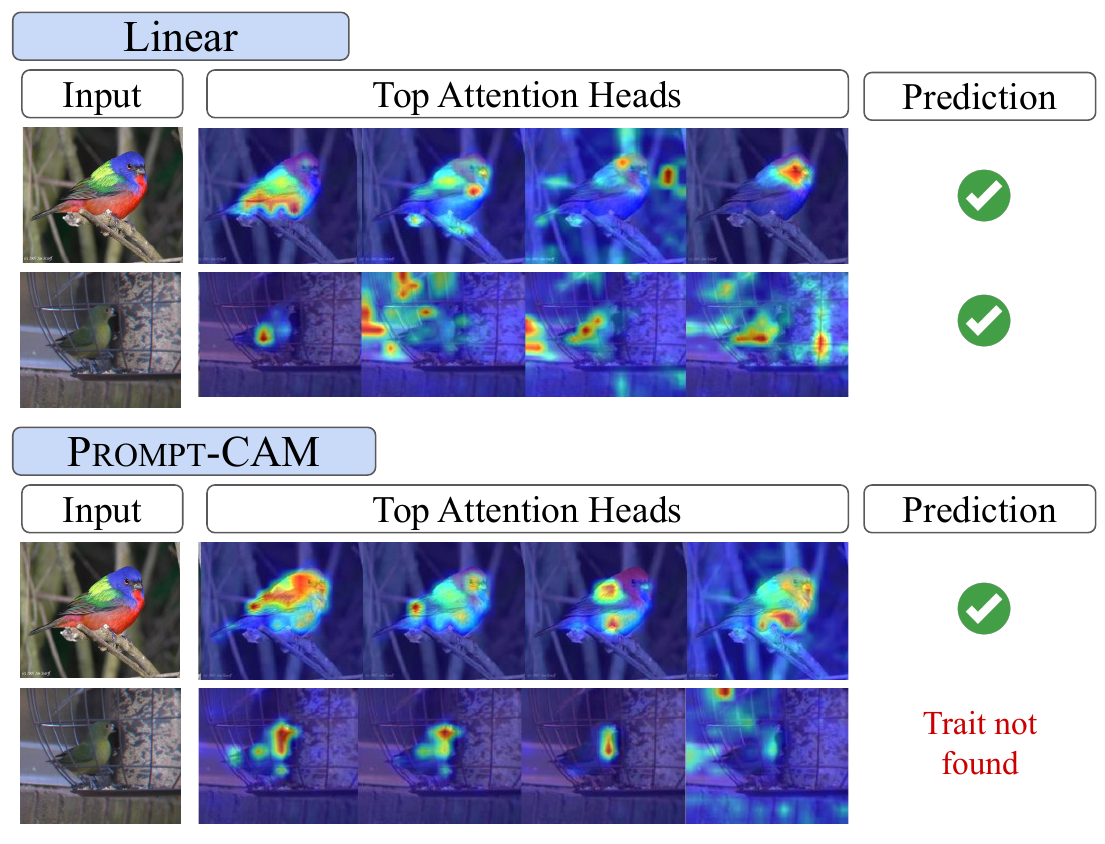}
    \caption{\textbf{Comparison of top attention heads for \Ours and Linear probing on two images of the species ``Painted Bunting.''} For the correctly classified image by both, \Ours focuses on meaningful traits such as the blue head, wings, tail, and red lower belly, while Linear probing produces noisy and less diverse heatmaps. For the other image, Linear probing relies on global memorized attributes for correct classification, whereas \Ours attempts to identify object-specific traits, resulting in an interpretable misclassification due to poor visibility of key features.}
    \label{supp_fig:linear_vs_pcam_misclassified}
    \vskip -8pt
\end{figure}

\begin{figure*}[t]
    \centering
    \includegraphics[width=\linewidth]{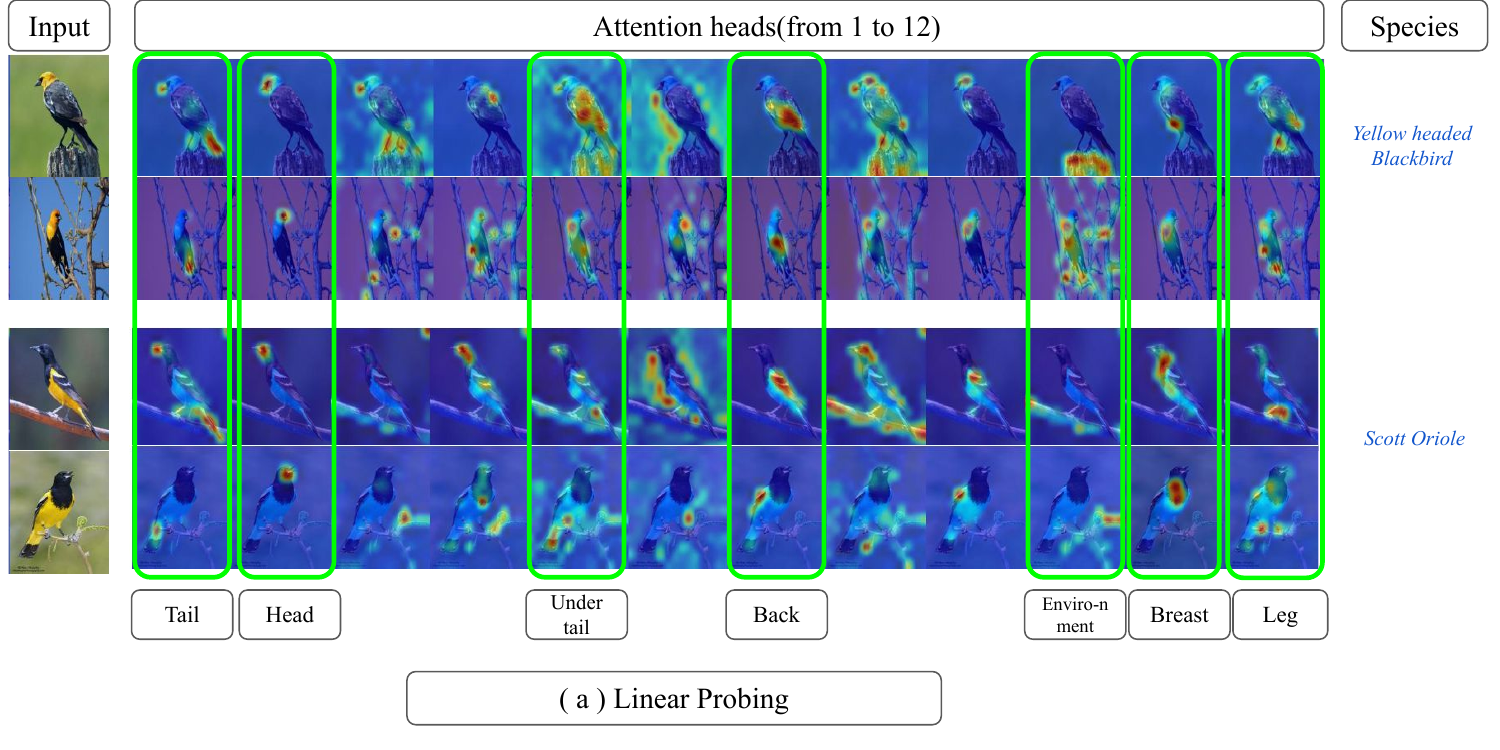}
    \includegraphics[width=\linewidth]{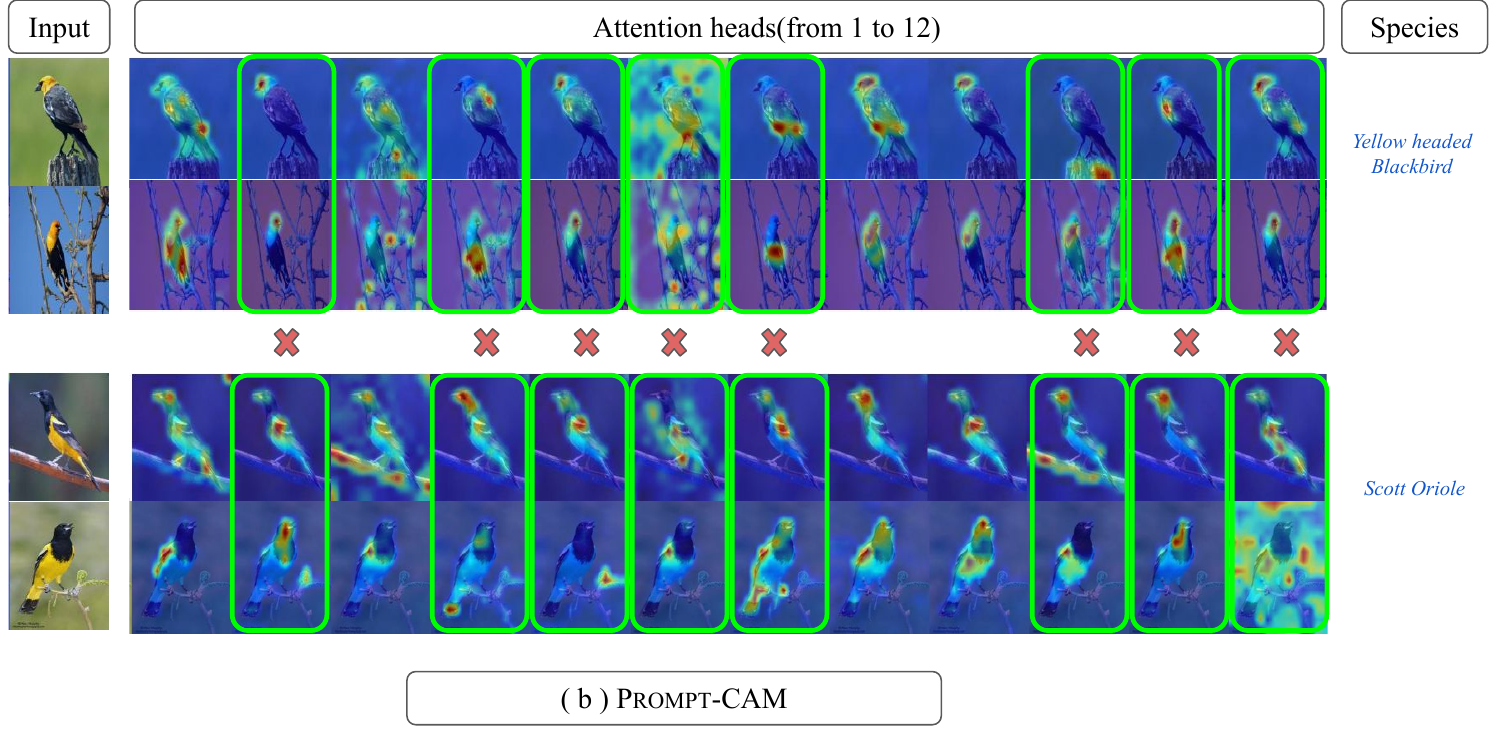}
    \caption{\textbf{Comparison of attention heatmaps for Linear Probing and \Ours.} On random images of ``Yellow Headed Blackbird'' and ``Scott Oriole'' from the CUB dataset, in (a), Linear Probing consistently focuses on similar body parts (e.g., tail, head, under-tail, wings) across all species, showing limited adaptability to traits specific to each class. In contrast, (b) \Ours (using pretrained DINO) dynamically adapts its attention to focus on distinct and meaningful traits required for class-specific identification. For instance, \Ours highlights traits such as the yellow head and breast for ``Yellow Headed Blackbird'' and the wing pattern for ``Scott Oriole''.}
    \label{supp_fig:linear_heatmap_for_three_species}
    \vskip -8pt
\end{figure*}

\mypara{Attention head vs species.}
In addition to image-level analysis, we conduct a species-level investigation to determine whether certain attention heads consistently focus on important traits across all images of a species. Using the greedy approach discussed in the above paragraph, we analyze each correctly classified image of a species \( c \) to iteratively select the attention head \( r^* \) that minimally impacts the probability of the correct class \( c \). 
We then examine how the probability \( s[c] \) changes as attention heads are progressively blurred or masked for all images of a species. This analysis, visualized in \autoref{supp_fig:head_vs_probablity}, demonstrates that for most species in the CUB dataset, approximately four attention heads capture traits critical for class prediction. In the \autoref{supp_fig:head_vs_probablity}, we highlight the top-2 attention heads for example images from various species. The results reveal that these heads consistently focus on important, distinctive traits for their respective species. For instance, in the case of the ``Cardinal", head-9 focuses on the black stripe near the beak, while head-10 attends to the red breast color—traits essential for identifying the species. Similarly, for ``Yellow-breasted Chat" and  ``Red-faced Cormorant", attention heads consistently highlight relevant features across their respective species. These findings emphasize the robustness of our approach in identifying class-specific discriminative traits and the flexibility of choosing any number of ranked important traits per species.

\begin{figure*}[t]
    \centering
    \includegraphics[width=0.9\linewidth]{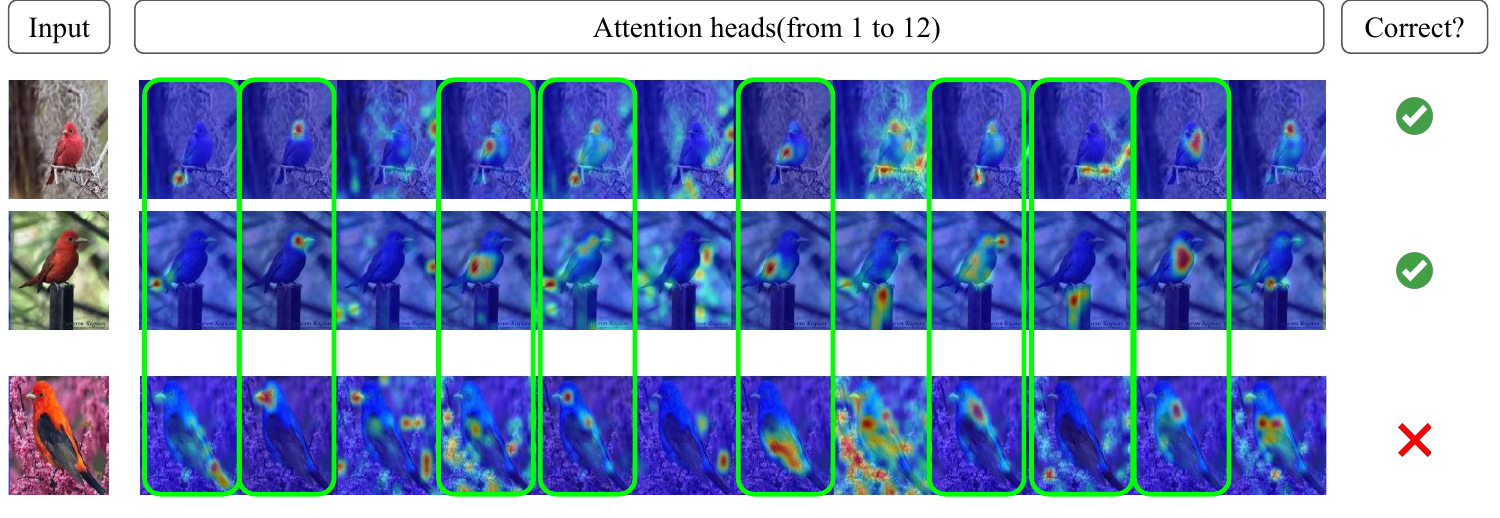}
    \caption{\textbf{Attention heatmaps of cls-token for Linear Probing on misclassified images.} For some random images of ``Scarlet Tanager" from the CUB dataset, Linear Probing highlights the same body parts across images, failing to provide meaningful insights into misclassifications. }
    \label{supp_fig:linear_heatmap_vs_pcam_12}
    \vskip -8pt
\end{figure*}

\section{Additional Experiment Settings}
\label{supp:exp_settings}
\subsection{Implementation Details}
\mypara{Dataset-specific settings.} For DINO backbone, the learning rate varied across datasets within the set $\{0.01, 0.1, 0.125\}$, selected based on dataset-specific characteristics. For Bird and MedLeaf, training was conducted for 30 epochs. For all other datasets, training was conducted for 100 epochs. For DINOv2 backbone, the learning rate varied across datasets within the set of $\{0.005, 0.01\}$, selected based on dataset-specific characteristics. For Insect, CUB, and Bird, training was conducted for 130 epochs. For all other datasets, training was conducted for 100 epochs. For DINOv2 backbone, the learning rate varied across datasets within the set of $\{0.05, 0.01\}$, selected based on dataset-specific characteristics. For all datasets, training was conducted for 100 epochs. A batch size of 64 was used for all datasets and all backbones.

\begin{figure}[t]
    \centering
    \includegraphics[width=1\linewidth]{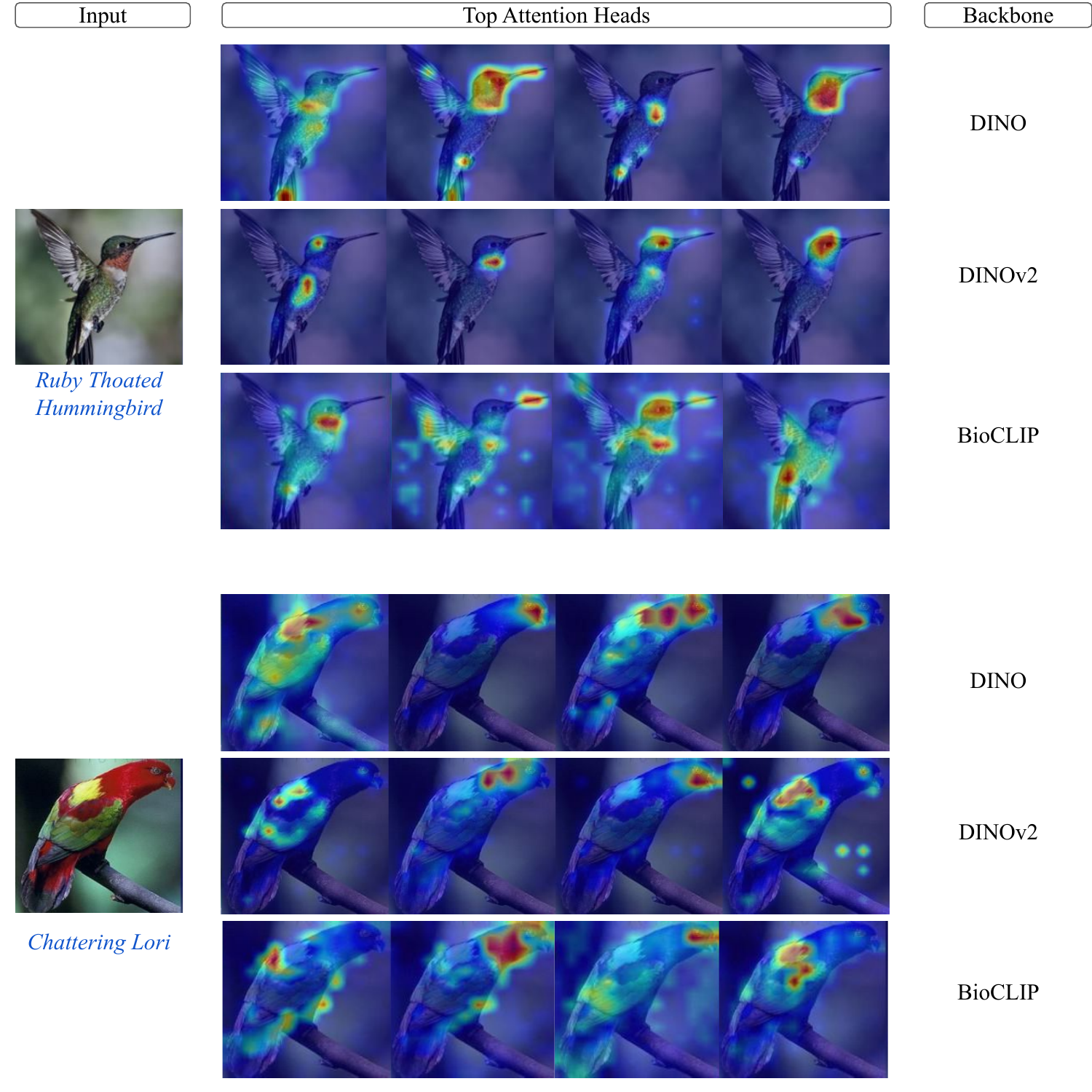}
    \caption{\textbf{Visualization of top attention heads of \Ours for DINO, DINOv2 and BioCLIP backbones.} For random correctly classified images from ``Ruby Throated Hummingbird'' and ``Chattering Lori'' species from Bird Dataset, top-4 attention heads (from left to right) are shown. \Ours can identify and locate meaningful important traits for species regardless of pre-trained visual backbone used. }
    \label{supp_fig:different_architecture_heatmap}
    \vskip -8pt
\end{figure}

\mypara{Optimization settings.} Stochastic Gradient Descent (SGD) optimizer with a momentum of 0.9. Weight Decay 0.0 was used for all datasets for DINO, 0.001 for the rest. A cosine learning rate scheduler was applied, with a warmup period of 10 epochs and cross-entropy loss was used.
\subsection{Baseline Methods}
We used \textbf{XAI methods} Grad-CAM, Score-CAM, and Eigen-CAM to compare \Ours performance with them on a quantitative scale. For qualitative comparison, we compare with a variety of \textbf{interpretable methods}, ProtoPFormer, TesNet, INTR, and ProtoPConcepts shown in \autoref{fig:interpretableComparison}.

\section{Additional Experiment Results}
\label{supp:exp_results}
\mypara{Model performance analysis.}
As discussed in \autoref{ss:vis}, we analyze misclassified examples by \Ours, illustrated in \autoref{fig:misclassified_images}. We attribute the slight decline in accuracy of \Ours to its approach of forcing prompts to focus on the object itself and its traits, rather than relying on surrounding context for classification.
In \autoref{supp_fig:linear_vs_pcam_misclassified}, we compare the heatmaps of two images of the species ``Painted Bunting". The first image, \(I_c\), is correctly classified by both \Ours and Linear probing, while the second image, \(I_m\), is correctly classified by Linear probing but misclassified by \Ours. The image \(I_m\) presents additional challenges: it is poorly lit, further from the camera, and depicts a less common gender of the species in the CUB dataset.

For \(I_c\), the top heatmaps from Linear probing appear noisy and less diverse compared to \Ours. In contrast, \Ours exhibits a more meaningful focus, with its top attention heads targeting the blue head, part of the wings, the tail, and the red lower belly—traits characteristic of the species.

In the case of \(I_m\), although Linear probing predicts the image correctly, its top attention heads fail to focus on consistent traits. Instead, they appear to rely on global features memorized from the training dataset, resulting in a lack of meaningful interpretation. On the other hand, \Ours, despite misclassifying \(I_m\), focuses its attention on traits within the object itself. The heatmaps reveal that \Ours attempts to identify relevant features, but the lack of visible traits in the image leads to an interpretable misclassification.

In \autoref{supp_fig:linear_heatmap_for_three_species}, the comparison between Linear Probing and \Ours in the attention heatmaps reveals a fundamental difference in their classification and trait identification approach. As shown in the heatmaps, Linear Probing uniformly distributes its attention across similar body parts, such as the tail, head, and wings, irrespective of the species being analyzed. This behavior indicates that Linear Probing relies on global patterns that may not be specific to any particular class. 
In contrast, for each species, \Ours focuses on specific traits important for differentiating one class from another. For example, in the case of the ``Yellow Headed Blackbird," \Ours emphasizes the yellow head and breast, traits unique to the species. Similarly, for the ``Scott Oriole," the yellow breast and wing patterns are prominently highlighted. By prioritizing traits essential for species identification, \Ours provides a more robust and meaningful framework for understanding model decisions.

Furthermore, in \autoref{supp_fig:linear_heatmap_vs_pcam_12}, we present attention heatmaps for random images of the ``Scarlet Tanager" species from the CUB dataset, generated using Linear Probing. Linear Probing consistently assigns attention to the same body parts (e.g., wings, head) across images, without providing meaningful insights into the reasons for misclassification. In contrast, \Ours (as shown in \autoref{fig:trait_manipulation} and \autoref{supp_fig:linear_vs_pcam_misclassified}) provides a more interpretable explanation for misclassifications. When \Ours misclassifies an image, it is evident that the misclassification occurs due to the absence of the necessary trait in the image, demonstrating its focus on biologically relevant and class-specific traits.

This analysis underscores \Ours prioritizes interpretability, ensuring that its classifications are based on meaningful and consistent traits, even at the cost of a slight accuracy decline.

\mypara{Human assessment of trait identification settings.} In \autoref{sub_sec:experiment_results}, we discussed how we measured robustness of \Ours with assessment from human observers.  To evaluate the effectiveness of trait identification, in the human assessment, we compared \Ours, TesNet \cite{wang2021interpretable}, and ProtoConcepts \cite{ma2024looks}. A total of 35 participants with no prior knowledge of the models participated in the study. Participants were presented with a set of top attention heatmaps (\Ours and INTR) or prototypes generated by each method and image-specific class attributes found in CUB dataset. Then they were asked to identify and check the traits they perceived as being highlighted in the heatmaps. The traits were taken from the CUB dataset, where image-specific traits are present. We used four random correctly classified images by every method, from four species ``Cardinal'', ``Painted Bunting'', ``Rose Breasted Grosbeak'' and ``Red faced Cormorant'' to generate attention heatmaps/prototypes.

The assessment revealed that participants recognized $60.49\%$ of the traits highlighted by \Ours, significantly outperforming TesNet and ProtoConcepts, which achieved recognition rates of $39.14\%$ and $30.39\%$, respectively. These findings demonstrate \Ours's superior ability to emphasize and communicate relevant traits effectively to human observers.

\begin{figure}[t]
    \centering
    \includegraphics[width=\linewidth]{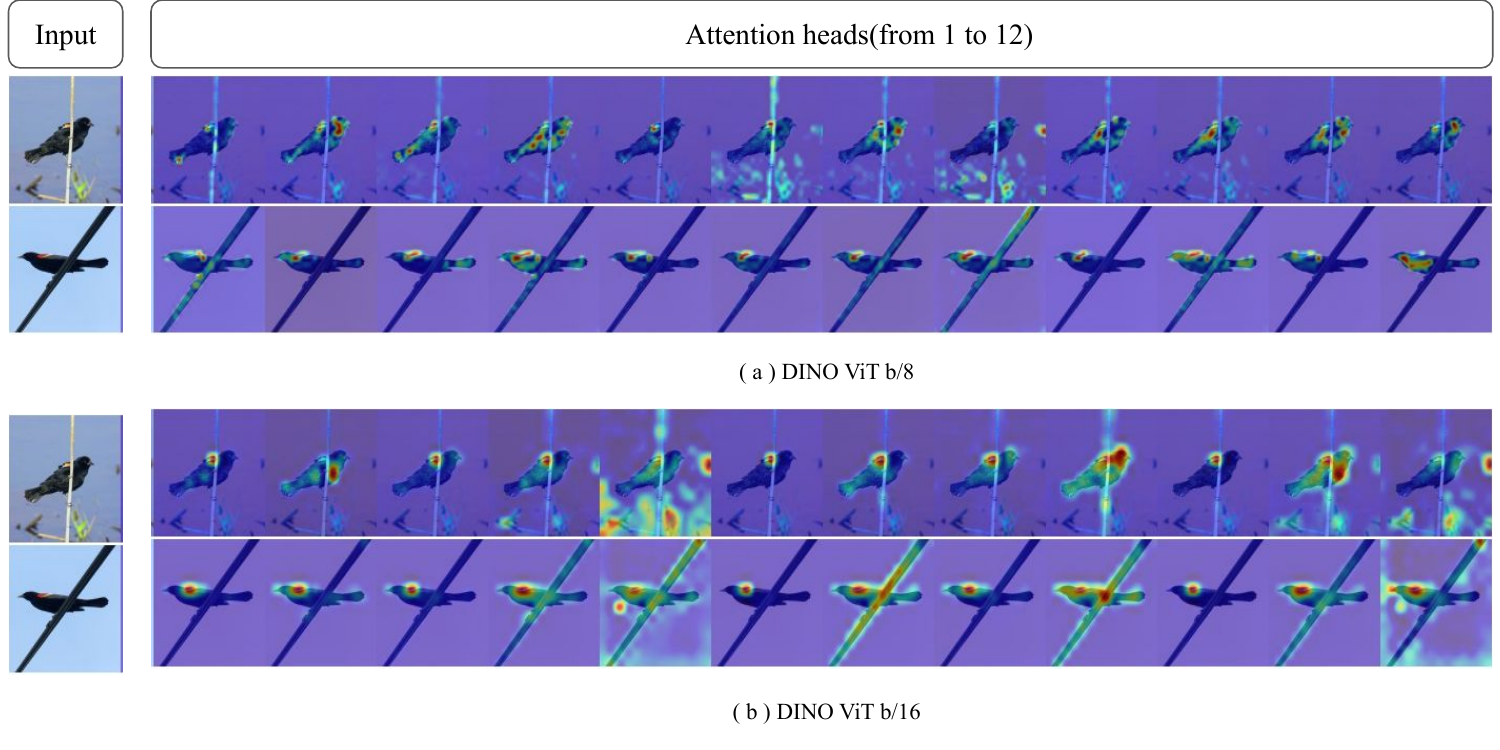}
    \caption{\textbf{Visualization of attention heads for pre-trained DINO backbone variants.} For correctly classified images of ``Red winged blackbird'', with \Ours, both DINO ViT b/16 and DINO ViT b/8 backbones can capture traits for classification.}\label{supp_fig:dino_v2_register_heatmaps}
    \vskip -8pt
\end{figure}

\mypara{\Ours on different backbones.}
We implement \Ours on multiple pre-trained vision transformers, including DINO, DINOv2, and Bioclip. In \autoref{sup_tab:accuracy_backbone}, we present the accuracy of \Ours across various datasets using different backbones: DINO (ViT-Base/16), DINOv2 (ViT-Base/14), and Bioclip (ViT-Base/16). For each model, we visualize the top-4 attention heads on the Bird Dataset in \autoref{supp_fig:different_architecture_heatmap}. Notably, Bioclip achieves higher accuracy on biology-specific datasets, which we attribute to its pre-training on an extensive biology-focused dataset, enabling it to develop a highly specialized feature space for these species. Additionally, we also evaluate \Ours on other DINO variations, ViT-Base/8 (accuracy: $73.9\%$) and ViT-Small/8 (accuracy: $68.3\%$) on the CUB dataset, achieving comparable performance and interpretability to DINO ViT-Base/16 (accuracy: $71.9\%$) (shown in \autoref{supp_fig:dino_v2_register_heatmaps}). This demonstrates \Ours's robustness, flexibility, and ease of implementation across various pre-trained vision transformer backbones and datasets.

\begin{table}[htb]
\centering
\caption{\textbf{Accuracy of \Ours on different backbones.} To show the flexibility and robustness, the accuracy of \Ours on multiple datasets is shown implemented on pre-trained vision transformers, DINO, DINOv2 and BioCLIP.}
\resizebox{\linewidth}{!}{
    \begin{tabular}{llcccccccc}
    \toprule
\multicolumn{2}{l}{}            & \multicolumn{1}{l}{Bird} & \multicolumn{1}{l}{CUB} & \multicolumn{1}{l}{Dog} & \multicolumn{1}{l}{Pet} & Insects-2 & Flowers & Med.Leaf & Rare Species \\
\midrule
\multirow{3}{*}{Ours} & DINO    & 98.2                     & 73.2                    & 81.1                    & 91.3                    & 64.7      & 86.4    & 99.1     & 60.8         \\
                      & DINOv2  & 98.2                     & 74.1                    & 81.3                    & 92.7                   & 70.6      & 91.9    & 99.6     & 62.2         \\
                      & BioCLIP & 98.6                     & 84.0                      & 73.1                    & 87.2                    & 71.8      & 95.7    & 99.6     & 67.1\\
                      \bottomrule
\end{tabular}
}
\label{sup_tab:accuracy_backbone}
\end{table}

\mypara{Taxonomical hierarchy trait discovery settings.}
In hierarchical taxonomic classification in biology, each level in the taxonomy leverages specific traits for classification. As we move down the taxonomic hierarchy, the traits become increasingly fine-grained. Motivated by this observation, we trained and visualized traits in a hierarchical taxonomic manner using the Fish Vista dataset.

We first constructed a taxonomic tree spanning from \textit{Kingdom} to \textit{Species}. For the \textit{Family} level, we aggregated all images belonging to the diverse species under their respective \textit{Family} and performed classification to assign images to the appropriate \textit{Family}. As shown in \autoref{fig:hieriarchial_trait}, even coarse traits, such as the tail and pelvic fin, were sufficient to classify an image of the species ``Amphiprion Melanopus'' to its' correct \textit{Family} (attribute information found in Fish Dataset).

At the \textit{Genus} level, we create a new dataset for each \textit{Family} by grouping all images from the children nodes of each \textit{Family} and dividing them into classes by their respective \textit{Genus}. For instance, within the ``Pomacentridae'' Family, finer traits like stripe patterns, pelvic fins, and tails became necessary to classify its' \textit{Genus} accurately for the same example image. Finally, at the \textit{Species} level, all images from the children nodes of each \textit{Genus} were used to create a new dataset and were divided into classes. For the example image in \autoref{fig:hieriarchial_trait}, distinguishing between these two species now requires looking at subtle differences such as the pelvic fin structure and the number of white stripes on the body for the same image from the ``Amphiprion Melanopus'' species. This hierarchical approach offers an exciting framework to discover traits in a manner that is both evolutionary and biologically meaningful, enabling a deeper understanding of trait importance across taxonomic levels.

\begin{figure*}[t]
    \centering
    \includegraphics[width=.8\linewidth]{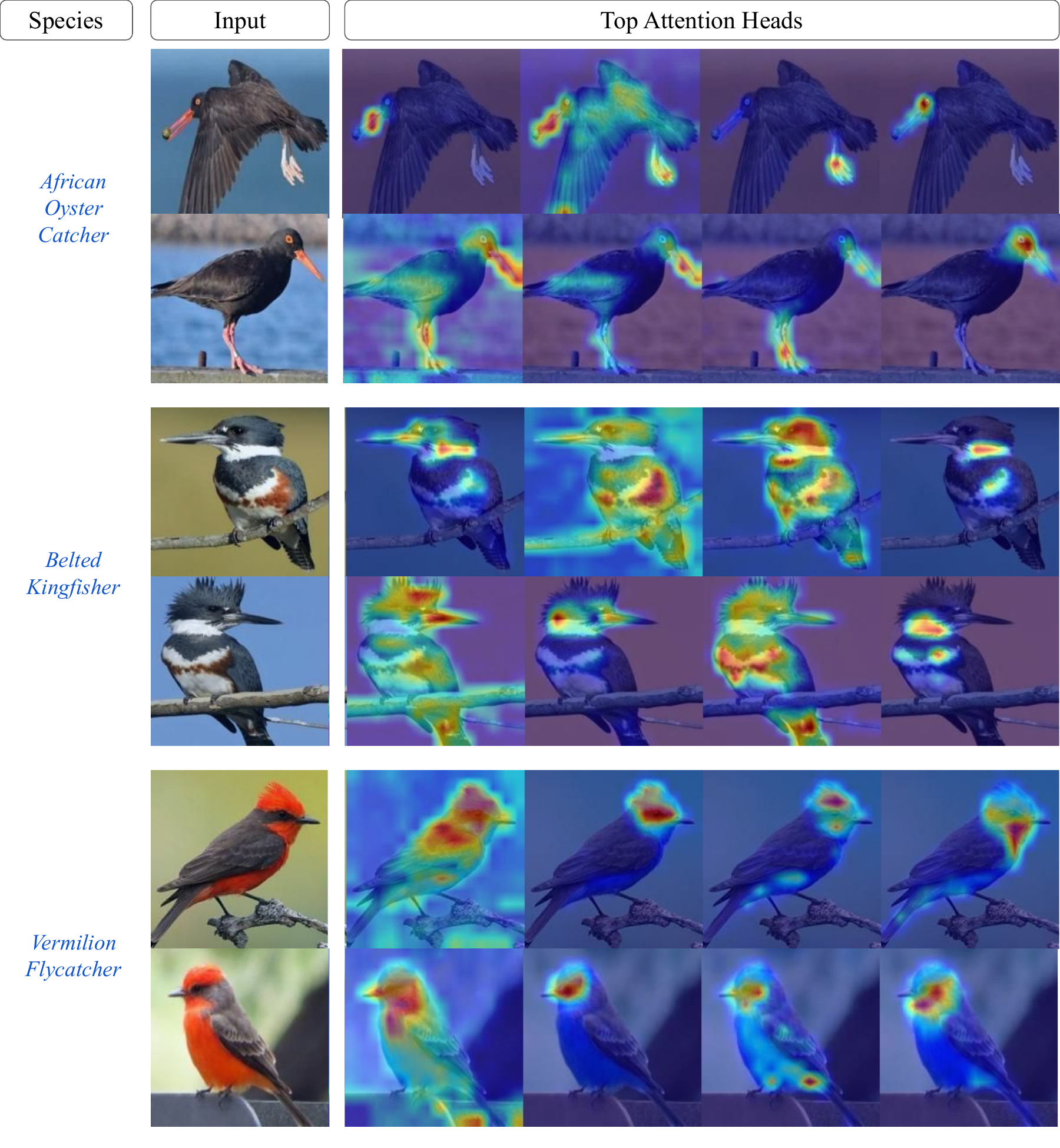}
    \caption{\small \textbf{Visualization of \Ours on Bird Dataset.} We show the top four attention maps (from left to right) per correctly classified test example, triggered by the ground-truth classes. As top head indices per image may vary, traits may not align across columns. }
    \label{supp_fig:more_vis_birds_1}
\end{figure*}

\begin{figure*}[t]
    \centering
    \includegraphics[width=.8\linewidth]{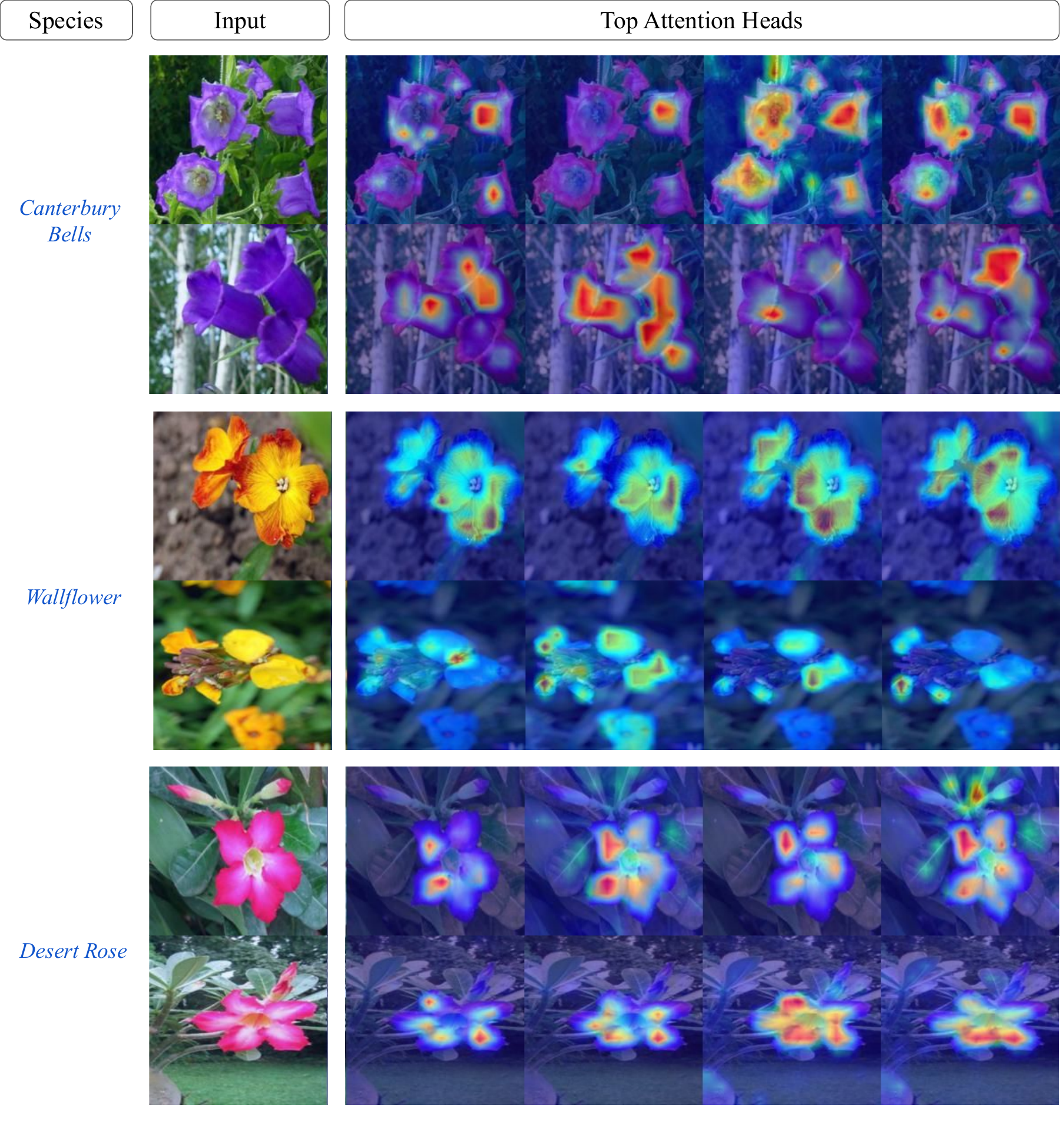}
    \caption{\small \textbf{Visualization of \Ours on Flower Dataset.} We show the top four attention maps (from left to right) per correctly classified test example, triggered by the ground-truth classes. As top head indices per image may vary, traits may not align across columns. }
    \label{supp_fig:more_vis_flow_2}
\end{figure*}

\begin{figure*}[t]
    \centering
    \includegraphics[width=.8\linewidth]{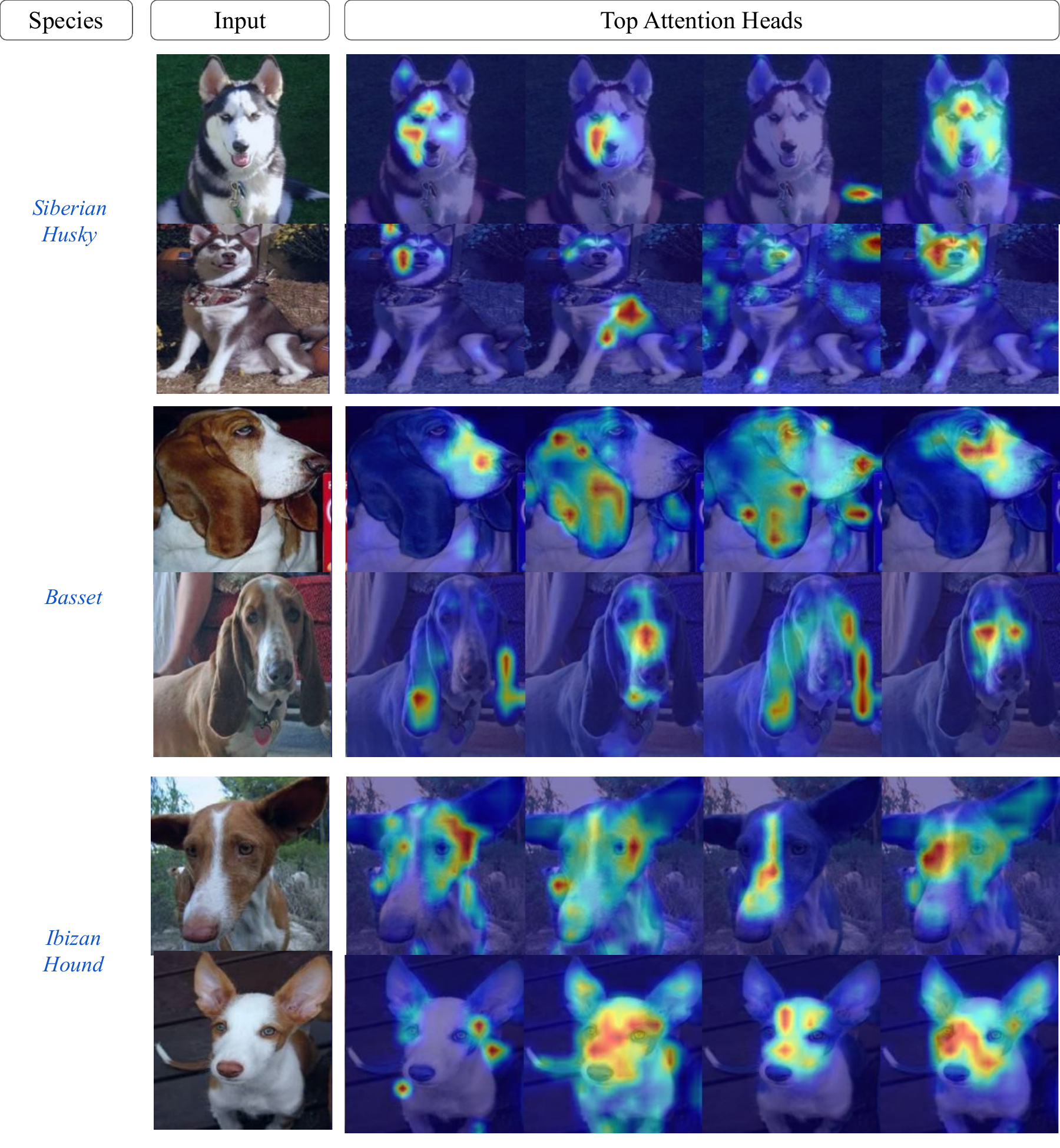}
    \caption{\small \textbf{Visualization of \Ours on Dog Dataset.} We show the top four attention maps (from left to right) per correctly classified test example, triggered by the ground-truth classes. As top head indices per image may vary, traits may not align across columns. }
    \label{supp_fig:more_vis_flow_3}
\end{figure*}

\section{More Visualizations}
\label{supp:more_visual}
In this section, we show the
top-4 attention maps triggered by ground truth classes for correctly predicted classes, for some datasets mentioned \autoref{supp:dataset}, following the same format of \autoref{fig: all_dataset_figure}. Each attention head of \Ours for each dataset
successfully identifies different and important attributes of each class of every dataset. For some datasets, if the images of a class are simple enough, we might need less than four heads to predict.

\end{document}